\documentclass[lettersize,journal]{IEEEtran}
\usepackage{amsmath,amsfonts}
\usepackage{algorithmic}
\usepackage{algorithm}
\usepackage{array}
\usepackage[caption=false,font=normalsize,labelfont=sf,textfont=sf]{subfig}
\usepackage{textcomp}
\usepackage{stfloats}
\usepackage{url}
\usepackage{verbatim}
\usepackage{graphicx}
\usepackage{cite}
\usepackage{booktabs}
\usepackage[hidelinks]{hyperref}
\usepackage{xcolor}
\usepackage{mathrsfs}
\usepackage{amsthm}

\theoremstyle{plain}

\theoremstyle{definition}

\theoremstyle{remark}
\newtheorem{remark}{Remark}

\begin{document}

\title{A Taxonomy of Structure from Motion Methods}

\author{
Federica Arrigoni
\thanks{\IEEEcompsocthanksitem F. Arrigoni is with the Dipartimento di Elettronica, Informazione e Bioingegneria (DEIB), Politecnico di Milano, Milano, Italy. \\ E-mail: federica.arrigoni@polimi.it}
}

\markboth{
}%
{Shell \MakeLowercase{\textit{et al.}}: A Sample Article Using IEEEtran.cls for IEEE Journals}


\maketitle

\begin{abstract}
Structure from Motion (SfM) refers to the problem of recovering both structure (i.e., 3D coordinates of points in the scene) and motion (i.e., camera matrices) starting from point correspondences in multiple images. It has attracted significant attention over the years, counting practical reconstruction pipelines as well as theoretical results.
This paper 
is conceived as a {conceptual} review of SfM methods, which are grouped into three main categories, according to which part of the problem -- between motion and structure -- they focus on. The proposed taxonomy brings a new perspective on existing SfM approaches as well as insights into open problems and possible future research directions. Particular emphasis is given on identifying the theoretical conditions that make SfM well posed, which depend on the problem formulation that is being considered. 
\end{abstract}


\section{Introduction}
\IEEEPARstart{L}{et} us consider a static scene in 3D space and suppose that the scene is captured by multiple cameras at different positions and viewing directions, thus producing a set of 2D images.
We are interested here in the \emph{inverse problem}, namely 3D reconstruction from images -- this is far from being trivial as there is loss of information when going from 3D to 2D, thereby motivating dedicated research. 
Typically, the starting point of a reconstruction pipeline is a set of matches, namely image points that correspond to the same (unknown) 3D point \cite{OzyesilVoroninskiAl17}. In this paper we assume that those correspondences are given.
Accordingly, the final 3D reconstruction is sparse (in the form of a 3D point cloud).
Cameras are sometimes referred to as ``views'' and the geometric structure of the scene/cameras as well as their relationships are studied under the term \textbf{multi-view geometry} \cite{HartleyZisserman04}, which sets the foundations of 3D reconstruction algorithms.
This problem is related to a wealth of applications including preservation of cultural heritage, medical imaging, virtual/augmented reality, novel view synthesis and autonomous driving, just to name a few.

Image-based 3D reconstruction typically involves two main sub-tasks, namely recovering a 3D representation of the scene (usually in the form of a 3D point cloud), as well as recovering the cameras that captured the input images.
The former is also called \emph{structure}, whereas the latter is also called \emph{motion}: this explains why the problem we are considering here is also referred to as \textbf{structure from motion (SfM)} in the literature \cite{OzyesilVoroninskiAl17}. We use this term since it is the most common in the literature, however, as will be clear in the sequel, ``structure from motion'' is appropriate only for a subset of methods that adopt a specific methodology.
SfM literature is extremely huge, ranging from efficient/robust pipelines (e.g., \cite{CrandallOwensAl11,ChatterjeeGovindu13,WilsonSnavely14,SenguptaAmirAl17,MagerandDel-Bue20,SarlinLindenbergerAl23}) to theoretical results (e.g., \cite{HartleyTrumpfAl13,NasihatkonHartleyAl15,WilsonBindel20,BratelundRydell23,ManamGovindu23}).
Observe that the underlying mathematical models of SfM have an explicit nature, which enhances theoretical understanding/interpretability while also fulfilling practical needs. 
Despite having a long history, SfM is still deemed relevant by the community and {actively researched nowadays}. 

\begin{figure}[t]
    \centering
    \includegraphics[width=0.95\linewidth]{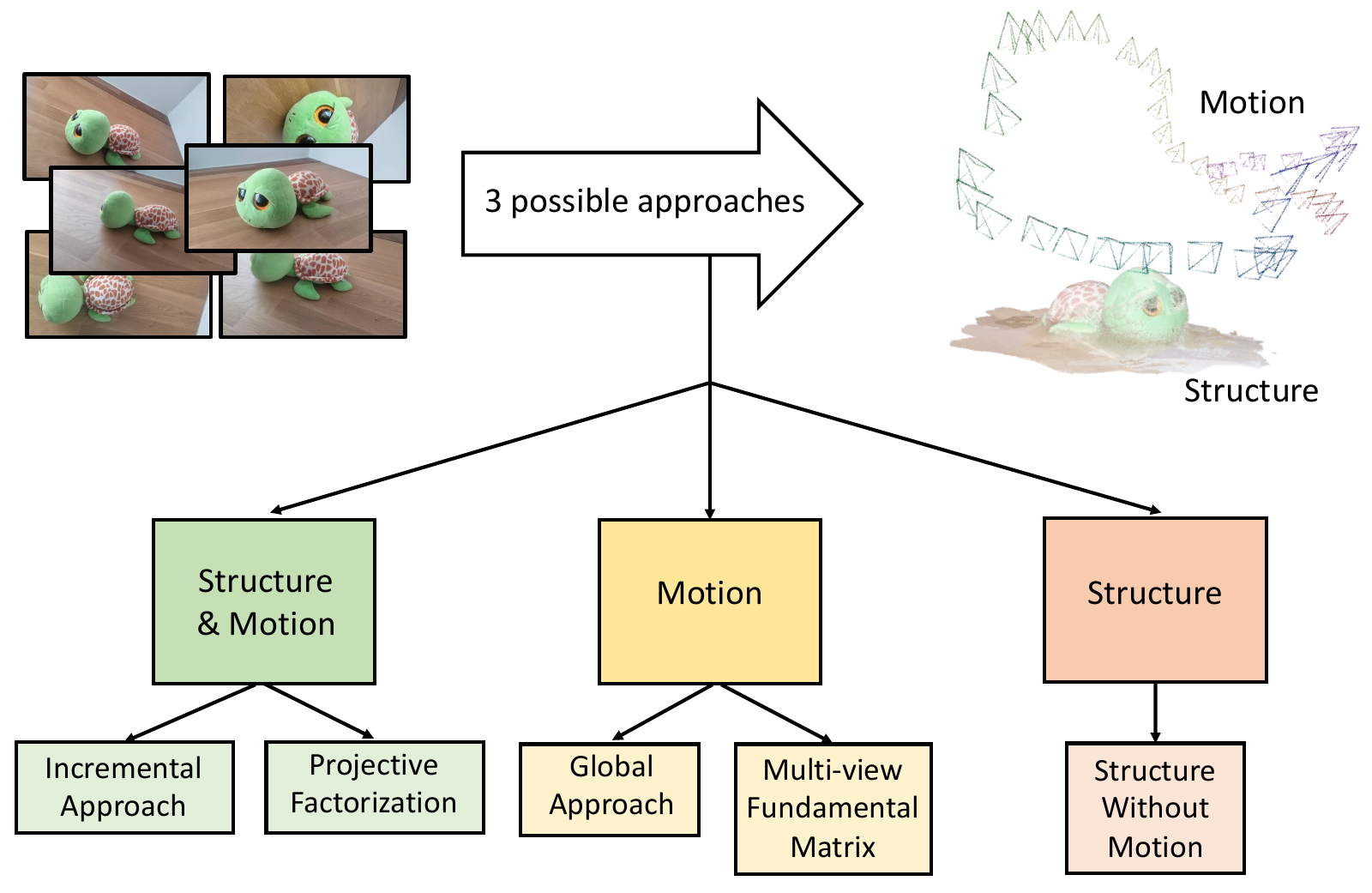}
    \caption{  The goal of Structure from Motion (SfM) is to recover both camera motion and scene structure starting from point correspondences in multiple images. According to the proposed taxonomy, methods can be divided into three categories, based on which part of the problem they focus on (motion, structure, motion and structure). Each category can be further divided into sub-categories based on the approach/assumptions, as described in the text.
    } 
    \label{fig:taxonomy}
\end{figure}

This paper is conceived as a \textbf{conceptual review} of SfM methods. Rather than following a historical perspective, a principled taxonomy is proposed (see Figure \ref{fig:taxonomy}) where existing approaches are grouped into three main categories, based on which part of the problem (structure and/or motion) they put attention on. This taxonomy brings a new perspective into SfM, giving rise to interesting insights and similarities between apparently disparate approaches. Particular emphasis is given to reviewing the \textbf{theoretical assumptions} that make SfM well posed, which typically depend on the specific formulation that is being considered. In this respect, graph representations of the SfM problem will be exploited to aid theoretical analyses. Although most methods in this area are traditional, in the sense that they are based on geometry and explicit optimization, we will also cover a few recent data-driven approaches for structure from motion.

An excellent survey on SfM was proposed eight years ago by Ozyesil et al.~\cite{OzyesilVoroninskiAl17}, which reviewed the status of knowledge up to that point, with specific focus on camera location recovery (in the calibrated case), scene structure estimation techniques, and the connection with the related problem of Simultaneous Localization and Mapping (SLAM). 
This paper differs from  \cite{OzyesilVoroninskiAl17} in several aspects: i) it follows a different conceptual flow based on a sound taxonomy; ii) it complements \cite{OzyesilVoroninskiAl17} with the most recent developments in the field, touching also deep learning approaches and new trends in the uncalibrated scenario (such as \cite{KastenGeifmanAl19}), that try to mimic the popular global SfM approach developed for calibrated cameras; iii) it gives significant emphasis to explain the theoretical assumptions and well-posedness implied by different SfM formulations.

The paper is organized as follows: Section \ref{sec_problem} formally defines the SfM problem, it introduces relevant notation and the proposed taxonomy made of three relevant categories; Section \ref{sec_SAM} is devoted to the first category of methods, that consider structure and motion simultaneously; Section \ref{sec_SFM} considers the second category of techniques, that focus on motion only; Section \ref{sec_SAM} describes the third category of approaches, that focus on structure instead; the conclusion is drawn in Section \ref{sec_conclusion} along with open issues in the field. This paper is inspired by recent tutorials on multi-view geometry  \cite{MagriArrigoni22,MagriArrigoni24}.

\section{Problem Formulation}
\label{sec_problem}

The goal of structure from motion (SfM) is to recover the following quantities, starting from multiple images:
\begin{itemize}
    \item \textbf{camera motion} = camera matrices;
    \item \textbf{scene structure} = coordinates of 3D points.
\end{itemize}
More precisely, the input is a set of correspondences across the images, that we assume given throughout this paper. In practice, such input correspondences are derived from the detection of repeatable keypoints across multiple views. See \cite{HeSunAl24} for an alternative detector-free approach.
We also assume that the underlying 3D scene is \textbf{static}, i.e., there are no moving objects. We refer the reader to \cite{SaputraMarkhamAl18} for a survey on dynamic structure from motion. In this section we formally define the SfM objective after introducing relevant notation (Section \ref{subsec_notation}). We also introduce the implications of the reconstruction ambiguity  (Section \ref{subsec_ambiguity}) and we explain the role of the so-called ``Bundle Adjustment'' (Section \ref{subsec_BA}). Finally, we introduce the proposed taxonomy (Section \ref{subsec_taxonomy}), which sets the basis for reviewing existing SfM approaches.

\subsection{Notation and Objective}
\label{subsec_notation}

Let $M_1, \dots, M_p$ denote the unknown coordinates of the 3D points present in the scene, represented as $4 \times 1$ vectors via homogeneous coordinates, where $p$ denote the total amount of points. Let $P_1, \dots, P_n$ denote the unknown matrices of the cameras, where $n$ denotes the total amount of cameras/images. In this paper we focus on the perspective camera model since it is widely used, considering both the calibrated and uncalibrated scenarios. In the \textbf{uncalibrated} case, the matrix of camera $i$ is represented as a $3 \times 4$ matrix of rank $3$ defined up to scale, whereas in the \textbf{calibrated} case it has this form:
\begin{equation}
    P_i = K_i [R_i \ | \ \mathbf{t}_i]
    \label{eq_calibrated_camera}
\end{equation}
where $K_i$ denotes the known calibration matrix, $R_i$ denotes the unknown $3 \times 3$ rotation matrix and $\mathbf{t}_i$ denotes the unknown $3 \times 1$ translation vector of the camera. The calibration matrix contains information about the intrinsic parameters of the camera (such as the focal length), whereas the rotation and translation encode the extrinsic parameters, defining the coordinate system of the camera (i.e., where it is located in space and how it is oriented). See \cite{HartleyZisserman04} for more details on different camera models. The center of the camera is denoted by $\mathbf{x}_i$ and can be retrieved via the following formula:
\begin{equation}
    \mathbf{x}_i = - R_i^{\mathsf{T}} \mathbf{t}_i.
    \label{eq_centre}
\end{equation}
Let $\mathbf{m}_{ij}$ denotes the known $3 \times 1$ vector representing the projection of 3D point $j$ onto image $i$, namely: 
\begin{equation}
    \mathbf{m}_{ij} \simeq P_i \mathbf{M}_j
\label{eq_projection}
\end{equation}
where $\simeq $ denotes equality up to scale.

Using the aforementioned notation, the SfM objective can be formalized as computing both camera matrices $P_i$ and scene points $\mathbf{M}_j$ starting from image correspondences $\mathbf{m}_{ij}$, such that Equation \eqref{eq_projection} is ``best'' satisfied for $i=1, \dots, n$ and $j=1, \dots, p$. A suitable optimization objective will be detailed in Section \ref{subsec_BA}. Recall that, in the calibrated case, the calibration matrices $K_i$ of the cameras are assumed known (which in turn can be retrieved, e.g., with external equipment like a checkerboard \cite{HartleyZisserman04}), therefore estimating cameras reduces to computing rotations and translations. In the uncalibrated case, the problem is also known as \emph{Projective SfM}.

\begin{remark}
Observe that Equation \eqref{eq_projection} defines an ill-posed problem in the case of a single image ($n=1$): in this scenario, there is no unique solution but there is an infinite number of 3D points (which defines an optical ray) that project to the same 2D image point. Single-view reconstruction becomes feasible when \emph{additional assumptions} are made, e.g., prior knowledge about the shape of objects \cite{FahimAminAl21}. Such a scenario, however, is not covered by this paper. Instead, we focus on reconstruction based on geometry and, accordingly, we assume that \textbf{multiple images} are given. We focus on the challenging $n>2$ case and do not analyze in detail the $n=2$ case.
\end{remark}

\subsection{Reconstruction Ambiguity and Degeneracies}
\label{subsec_ambiguity}

A relevant theoretical question is whether the solution to Equation \eqref{eq_projection} is \textbf{unique}. Actually, the answer is negative due to the following straightforward computations:
\begin{equation}
    \mathbf{m}_{ij} \simeq P_i \underbrace{Q Q^{-1}}_{I_4} \mathbf{M}_j = (P_i Q) (Q^{-1} M_j)
\label{eq_projection_ambiguity}
\end{equation}
where $Q$ denotes any $4 \times 4$ invertible matrix and $I_4$ is the $4 \times 4$ identity matrix. The effect of the $Q$ matrix is changing the global coordinate system. The above equations means that, if we right multiply all the cameras by the \emph{same} invertible matrix (and, accordingly, we left multiply all the 3D points by the inverse of such a matrix), then we get another valid solution, i.e., another set of cameras and 3D points that project onto the same image points. This means that the solution to SfM is not unique but it is defined up to a global \textbf{projective transformation}, in
the uncalibrated case, which represents the most general scenario where no assumptions are made. In this case, the final reconstruction is called \emph{projective}: it can be subject to distortions with respect to the real Euclidean scene, due to the fact that (e.g.) projective transformations do not preserve parallelism but incidence relations. 

In the calibrated case, it can be checked that the solution to SfM is defined up to a global rotation, translation and scale, which define a \textbf{similarity transformation} \cite{HartleyZisserman04}. In this scenario, the final reconstruction is called \emph{metric} or \emph{Euclidean} in the sense that it represents a scaled/rotated/translated version of the real scene, where (e.g.) angles are preserved. Note that the true scale of the scene can not be determined without extra information.
Under suitable assumptions, it is possible to upgrade an uncalibrated/projective reconstruction to a calibrated/Euclidean one --
this process is also known as the \emph{self-calibration} problem (see \cite{HartleyZisserman04} for details).

\begin{remark}
Observe that a global projective (respectively, similarity) transformation represents the \emph{minimum/ideal} amount of ambiguities inherent in an uncalibrated (respectively, calibrated) SfM problem. 
In principle, it is possible that the ambiguities are described by multiple transformations (instead of a single one): in this case, the problem instance is called \textbf{degenerate} \cite{HartleyZisserman04}.
In other terms, in a degenerate case, there are multiple solutions to the SfM task, which can not be reduced to a single one modulo a global transformation, therefore the problem instance is ill posed and no method will produce a useful reconstruction. 
Therefore, it is important to know under which conditions a problem is well-posed, or, in other terms, in which situations a degeneracy appear.
While a complete taxonomy of degeneracies is available for the case of two views \cite{HartleyZisserman04,Bratelund24b,KahlHartley02}, the scenario of multiple images still presents open issues, as also mentioned in a recent survey on Algebraic Vision \cite{KileelKohn23}.
Most theoretical studies on degeneracies refer to specific formulations of SfM or make additional assumptions, as will be detailed in the sequel.  
\label{remark_degeneracy}
\end{remark}

\subsection{Bundle Adjustment}
\label{subsec_BA}

We now  move to the practical task of {how} to solve SfM. 
A reasonable approach is to minimize the \textbf{reprojection error}, that is the distance between the left and right sides in
Equation~\eqref{eq_projection}: the former represents the input image points, whereas the latter represents new image points obtained by reprojecting the estimated 3D points via the found cameras. By summing over all points/cameras, this results in the following problem:
\begin{equation}
 \min_{\substack{P_1, \dots, P_n \\ M_1, \dots, M_p}} \sum_{i=1}^n \sum_{j=1}^p  d( \mathbf{m}_{ij} , P_i \mathbf{M}_j   )^2
\end{equation}
where $d(\cdot , \cdot)$ denotes a proper distance in the image plane. 
This process is also known as \emph{Bundle Adjustment} \cite{TriggsMcLauchlanAl00} because it can be interpreted as adjusting the ``bundle of rays'' between each camera centre and the set of 3D points (or, equivalently, between each 3D point and the set of camera centres).
To cope with missing/wrong correspondences, a more general cost function can be considered:
\begin{equation}
     \min_{\substack{P_1, \dots, P_n \\ M_1, \dots, M_p}} \sum_{i=1}^n \sum_{j=1}^p w_{ij} \rho \Big( d( \mathbf{m}_{ij} , P_i \mathbf{M}_j   ) \Big)
     \label{eq_BA}
\end{equation}
where $\rho$ denotes a robust loss (such as the L1 loss or the Cauchy loss \cite{HollandWelsch77}) and $w_{ij}$ is an indicator variable:
\begin{equation}
    w_{ij} =
    \begin{cases}
        1 \quad \text{if point $j$ is visible in image $i$} \\
        0 \quad \text{otherwise.}
    \end{cases}
\end{equation}

Equation \eqref{eq_BA} defines a non-linear optimization problem, which does not admit closed-form solutions. Accordingly, the unknown structure and motion parameters are recovered with an \textbf{iterative method} \cite{TriggsMcLauchlanAl00}. The most popular choice is the Levenberg-Marquardt algorithm, that is a combination of Gradient Descent -- from which it inherits a rapid decrease in the cost function
-- and Gauss-Newton -- from which it inherits a rapid convergence in the neighborhood of the solution.
In this process, the derivatives of the objective function with respect to the parameters must be computed, which are typically collected in a Jacobian matrix. A simplified formulation can be derived, by exploiting the fact that such Jacobian matrix has a sparse structure: each term in the cost involves only one camera/point; each point may be visible only in a subset of cameras. Although these observations can boost efficiency, still, Equation \eqref{eq_BA} defines an extremely large minimization problem, due to the number of parameters involved.
In this respect, several efforts have been made in recent years to improve the efficiency of Bundle Adjustment, by reasoning on partitioned/distributed optimization \cite{ZhangZhuAl17,LeiZixinAl20,RenLiangAl22,ZhengChenAl23}, novel computer architectures \cite{OrtizPupilliAl20}, QR decomposition \cite{DemmelSommerAl21} and power series approximation of the matrix inverse \cite{WeberDemmelAl23}.

The main drawback of bundle adjustment is that it requires to be initialized close to the solution in order to work well in practice, otherwise it may converge to a local minimum. For this reason, bundle adjustment typically does not solve SfM alone but it serves as \textbf{final refinement}. This justifies that a wide spectrum of methods were developed over the years to find a good initialization for bundle adjustment -- they will be reviewed in the sequel. Alternative approaches were proposed in the literature to refine structure and motion, such as a featuremetric refinement \cite{SarlinLindenbergerAl23}, which aligns sparse and deep features to get better correspondences,  essential for the success of SfM. 
Very few works concentrate on initialization-free bundle adjustment \cite{HongZachAl16,ZachHong18,IglesiasNilssonAl23,WeberHongAl24}, which is still considered an open problem in the field. 
The main idea behind this research line is to exploit the Variable Projection Method, which eliminates a subset of the unknowns, and it is known to have a wider basin of convergence.

\subsection{Proposed Taxonomy}
\label{subsec_taxonomy}

The fact that bundle adjustment typically requires good starting values for structure and motion, it motivated extensive research to provide such initialization, as anticipated in Section \ref{subsec_BA}. 
For the sake of providing a conceptual review, we group existing approaches into three main categories\footnote{Other taxonomies are possible as well.}, according to which part of the problem they focus on:
\begin{itemize}
    \item \textbf{Structure and Motion}, where structure and motion are considered simultaneously -- reviewed in Section \ref{sec_SAM};
    \item \textbf{Structure from Motion}, where the focus is on motion (and structure is derived subsequently via standard tools) -- reviewed in Section \ref{sec_SFM};
    \item \textbf{Structure without Motion}, where the focus is on structure (and motion can be possibly derived subsequently via standard tools) -- reviewed in Section \ref{sec_SWM}.
\end{itemize}
The proposed taxonomy is illustrated in Figure \ref{fig:taxonomy} and will be presented in depth in the coming sections.  
Considering the amount of SfM papers\footnote{Note that the term ``structure from motion'' appears about 200K times by searching on Google Scholar.}, citing all of them would be nearly impossible, therefore -- for each category -- we only describe a few representative approaches, with a focus on the most recent ones. Specifically, we explain the general methodology (without giving all the implementation details of individual algorithms) and also touch theoretical considerations.

\begin{figure}[t]
    \centering  \includegraphics[width=0.9\linewidth]{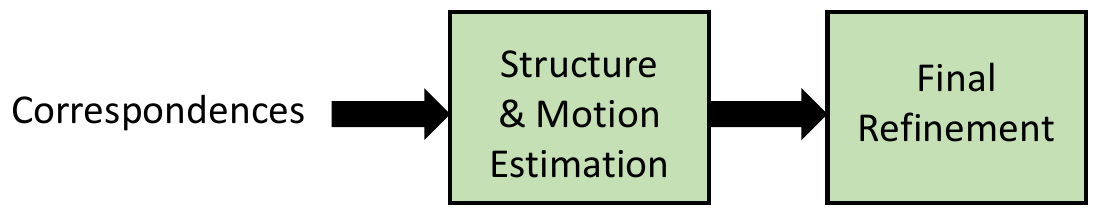}
    \caption{ The first category in the proposed SfM taxonomy recovers structure and motion simultaneously, starting from point correspondences. A final refinement (bundle adjustment) is applied at the end. Sequential/Hierarchical SfM and the projecive factorization belong to this category.
    } 
    \label{fig:SAM}
\end{figure}

\section{Structure and Motion}
\label{sec_SAM}

The first category in the proposed taxonomy comprises methods that 
address structure and motion estimation in a \emph{joint} manner, as summarized in Figure \ref{fig:SAM}. The usual starting point of SfM is a set of correspondences and a final refinement is applied at the end to improve the quality of the reconstruction.
The sequential approach (Section \ref{subsec_sequential}) and projective factorization (Section \ref{subsec_factorization}) belong to this category. Theoretical conditions of these approaches are discussed in Section \ref{subsec_SAM_theory}.

\subsection{Sequential Approach}
\label{subsec_sequential}

\begin{figure}[t]
    \centering  \includegraphics[width=0.75\linewidth]{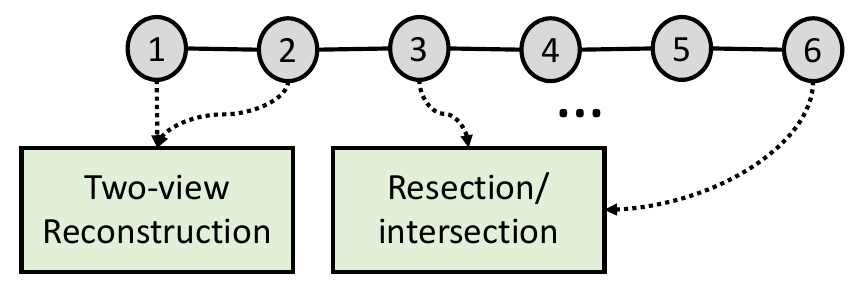}
    \caption{Sequential SfM belongs to the first category in the proposed taxonomy, which considers structure and motion in a joint manner. Assuming that images are organized in a sequence, two-view reconstruction is first performed; then, the remaining images are progressively included by alternating resection and intersection, thereby estimating new cameras and 3D points.
    } 
    \label{fig:SAM_sequential}
\end{figure}

\begin{figure*}
    \centering  
    \subfloat[Sequence]{
    \label{subfig:sequence}
    \includegraphics[height=0.2\linewidth]{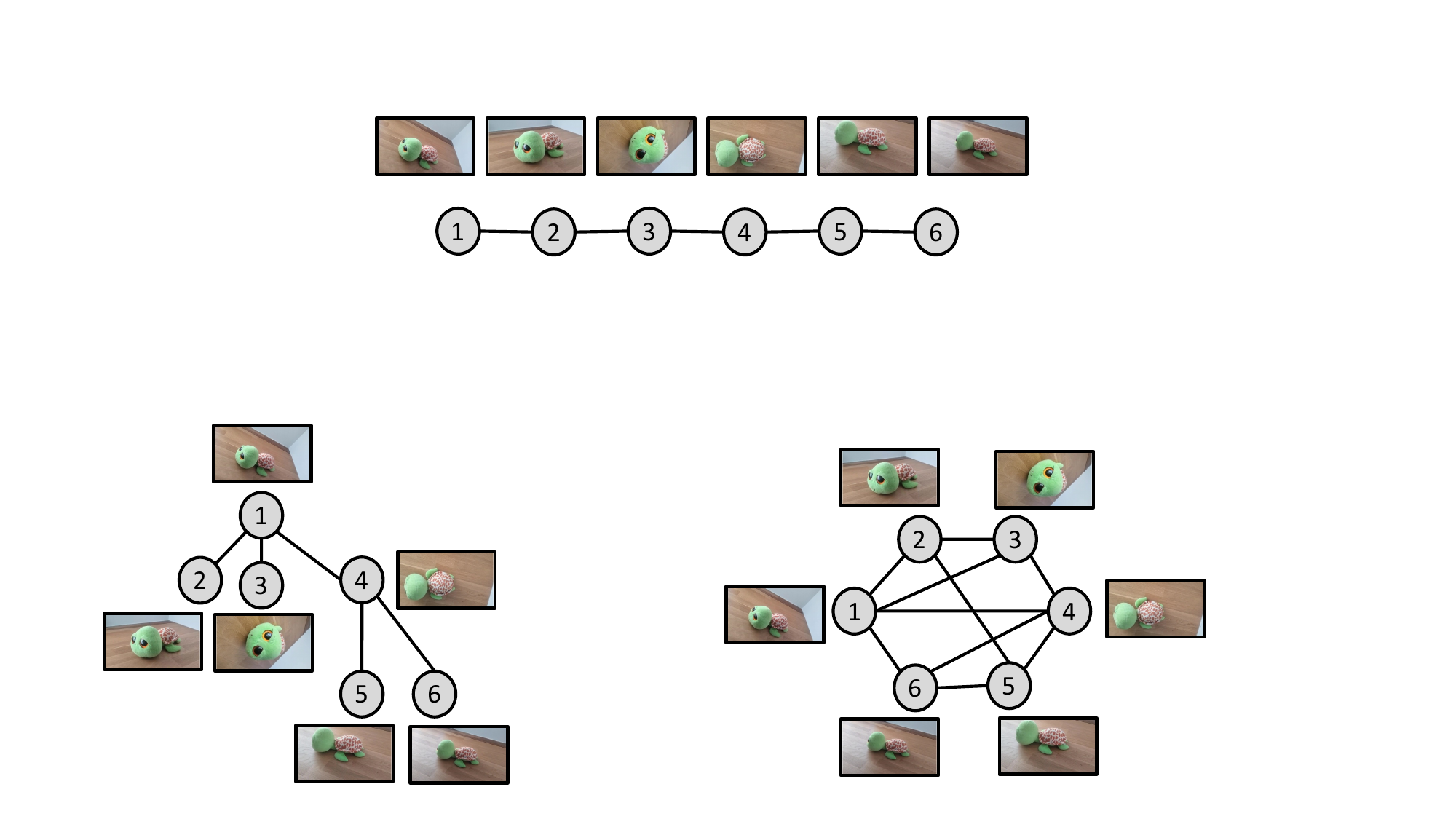} }
    \hfill 
    \subfloat[Tree]{
    \label{subfig:tree}
    \includegraphics[height=0.2\linewidth]{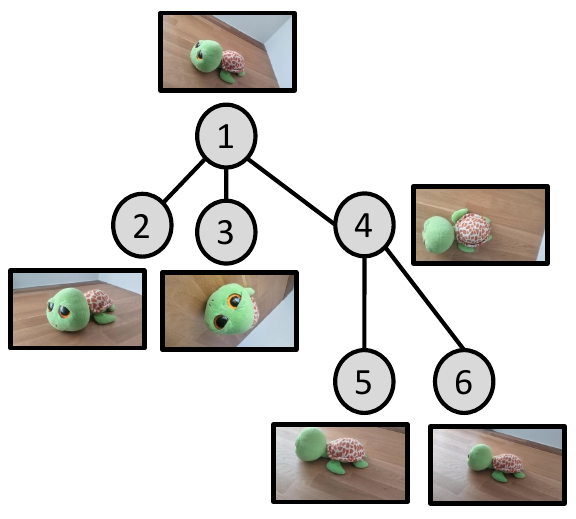} }
    \hfill 
    \subfloat[Viewing graph]{
    \label{subfig:viewing_graph}
    \includegraphics[height=0.2\linewidth]{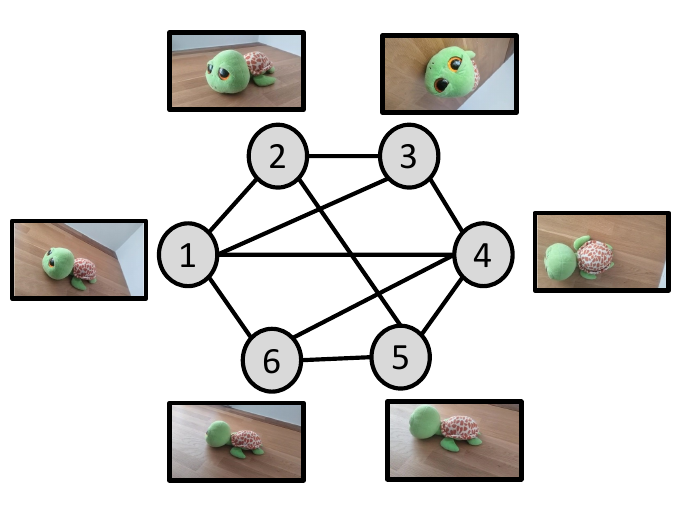} }
    \caption{There exist several ways to organize images in convenient abstract structures. In a {sequence}, images are consecutively captured, as happens (e.g.) in a video. In a tree, there are hierarchical relationships between overlapping images, where each node/image is connected with the parent node. In a viewing graph, each node corresponds to an image and overlapping images are put in relation via edges -- this is suitable for unordered image collections (e.g., taken from the internet). Note that we can view the sequence and the tree as special cases of the viewing graph, which is the most general representation.
    } 
    \label{fig:graphs}
\end{figure*}

One of the most popular approaches to recover structure and motion is
\textbf{Sequential SfM} (also known as {Incremental} SfM) \cite{SnavelySeitzAl06,AgarwalSnavelyAl09,FrahmFite-GeorgelAl10,Wu13,SchonbergerFrahm16}. Assuming that images are organized in a sequence (see Figure \ref{subfig:sequence}), the idea is to progressively build an increasing model composed of cameras and 3D points. Specifically, as shown in Figure \ref{fig:SAM_sequential}, two main steps are implied:
\begin{enumerate}
    \item Initializing the model, based on an initial image pair; 
    \item Updating the model, each time a new image is processed, until there are no more images left.
\end{enumerate}
Step 1 is a standard two-view reconstruction problem, which can be addressed as follows. Starting from correspondences in the first two images in the sequence, the \emph{essential matrix} (calibrated case) or the \emph{fundamental matrix} (uncalibrated case) of the pair can be recovered  -- a plenty of algorithms (including minimal solvers) are available for this task \cite{HartleyZisserman04}. In the calibrated scenario, the Singular Value Decomposition (SVD) of such essential matrix (denoted by $E_{12}$) permits to retrieve the relative rotation (denoted by $R_{12}$) and relative translation (denoted by $\mathbf{t}_{12}$) between the two cameras, which in turn are used to instantiate camera motion as follows:
\begin{equation}
    P_1 = K_1 [I_3 \ | \ \mathbf{0}], \quad 
     P_2 = K_2 [R_{12} \ | \  \mathbf{t}_{12}].
\end{equation}
Note that these two camera matrices are compliant with Equation \eqref{eq_calibrated_camera}.
The rotation matrix and translation vector of the first camera are set to the identity and zero, respectively -- this means that the world reference frame is aligned with the first camera, fixing the global rotation/translation ambiguity. Note that the relative translation can only be determined up to scale from the essential matrix \cite{HartleyZisserman04}. It is typically normalized to unit norm (i.e., $ || \mathbf{t}_{12} || = 1 $), thereby fixing the global scale ambiguity. 
Also in the uncalibrated case it is possible to recover camera motion in closed-form \cite{HartleyZisserman04}, exploiting the following formulas:
\begin{equation}
    P_1 = [I_3 \ | \ \mathbf{0}], \quad 
    P_2 = [ [\mathbf{e}]_{\times} F_{12}  + \mathbf{e} \mathbf{v}^{\mathsf{T}} \ | \  \lambda \mathbf{e} ]
\end{equation}
where $ \mathbf{e} $ denotes the epipole satisfying $\mathbf{e}^{\mathsf{T}} F_{12}=\mathbf{0}$, $[\mathbf{e}]_{\times} $ denotes the skew-symmetric matrix associated with the cross product with $\mathbf{e}$, $\mathbf{v}$ is an arbitrary 3-vector, and $\lambda$ is any non-zero scalar. Fixing the first camera to the world reference system and choosing specific values for $\mathbf{v}$, it corresponds to fixing the global projective ambiguity. 
Once the two camera matrices have been fully determined, the 3D points can be reconstructed by solving a \emph{triangulation} problem (also known as intersection), where a plenty of algorithms is available, including linear least squares \cite{HartleyZisserman04}. 
See also Table \ref{tab_resection}.

\begin{table}[t]
\centering
\caption{ 
Relevant problems in multi-view geometry.
 \label{tab_resection}}
{\small
\begin{tabular}{c c c} 
\toprule
  & Structure & Motion \\
\midrule
Triangulation / intersection & Unknown & Known  \\
Resection / PnP & Known & Unknown \\
SfM / reconstruction & Unknown & Unknown \\
\bottomrule
\end{tabular}
}
\end{table}

After model initialization (Step 1), structure and motion are available for the first two cameras, constituting an initial reconstruction. In order to update the model when adding a new image (Step 2), the available image correspondences are exploited: \emph{thanks to the initial reconstruction}, those 2D correspondences automatically translate into a set of 2D-3D correspondences for the third camera, which is therefore computed by solving a \emph{resection} problem \cite{HartleyZisserman04}. In the calibrated case, this single-view process is also known as \emph{perspective-n-point (PnP)}. See also Table \ref{tab_resection}.
The structure is then updated via triangulation, starting from the matrices of these three cameras and input 2D correspondences:
the position of 3D points that are observed in the new image is refined;
a new 3D point is initialized when we have a correspondence not related to an existing point.
This procedure is re-applied to all the images sequentially. Note that this approach naturally deals with missing data, i.e., it does not require the same points to be visible in all the images.

\begin{remark} 
The sequential SfM pipeline  is also called ``resection-intersection'', since it alternates between adding new cameras (by resection) and adding new points (by intersection).
Recall that SfM as a whole is more difficult than these tasks (resection and intersection) since both cameras and 3D points are unknown (see Table \ref{tab_resection}). In this respect, the idea of the sequential approach is to split the SfM computation into sub-problems that are easier to address. 
Observe that this method is not global, so it may suffer from error accumulation. For this reason, frequent bundle adjustment is needed to contain error propagation. Another disadvantage is that the choice of the initial pair may be critical.
Notwithstanding these considerations, sequential SfM currently represents the most successful pipeline in real scenarios. 
\end{remark}

There exist several systems that work well in practice.
One of the most popular is the COLMAP library\footnote{The COLMAP SfM software library received the PAMI Mark Everingham Prize in 2020. This prize is typically awarded for a selfless contribution of significant benefit to other members of the Computer Vision community.
} \cite{SchonbergerFrahm16}, which is based on careful implementation choices. Some examples are:
\begin{itemize}
    \item \emph{Next best view selection} -- it tends to favour images with more visible points and more uniform distribution.
    \item \emph{Local Bundle Adjustment} -- it is applied on the set of most connected images. 
    \item \emph{Redundant view mining} -- it clusters redundant cameras, which are then collapsed into a single camera.
\end{itemize}

\begin{remark}
Incremental reconstruction can be generalized to the case where images are organized in a tree instead of a sequence \cite{ToldoGherardiAl15,NiDellaert12}, as shown in Figure \ref{subfig:tree}. In this scenario (also called \textbf{Hierarchical SfM}), the procedure is as follows. First, many two-view problems are solved at the leaves of the tree (similarly to model initialization of the sequential approach). Then, the tree is traversed from the leaves to the root, and one of these operations takes place at each node:
\begin{itemize}
    \item one image is added (by resection/intersection, similarly to the model update of the sequential approach);
    \item two independent 3D reconstructions are merged -- by reasoning in terms of the structure, this process can be viewed as a 3D registration problem \cite{BeslMcKay92}.
\end{itemize}
\label{remark_hierarchical}
\end{remark}

\subsection{Projective Factorization}
\label{subsec_factorization}

\textbf{Projective Factorization} 
represents one of the oldest SfM approaches working with uncalibrated cameras \cite{SturmTriggs96}. It is an elegant method that recovers both structure and motion in a single step, under suitable assumptions. The main idea is that Equation \eqref{eq_projection} -- which formally defines the projection of 3D points via cameras onto images -- can be equivalently expressed as an equality by introducing explicit scales:
\begin{equation}
     \mathbf{m}_{ij} \simeq P_i \mathbf{M}_j \ \Longleftrightarrow \ 
     d_{ij} \mathbf{m}_{ij} = P_i \mathbf{M}_j .
     \label{eq_projection_depths}
\end{equation}
Such scales are called \emph{projective depths} and they are typically unknown. For simplicity of exposition, we first consider the case where all projective depths are given and all image points are visible in all the images (i.e., \emph{full visibility}). We will discuss the cases of partial visibility and unknown depths later.

\begin{figure}[t]
    \centering  \includegraphics[width=1\linewidth]{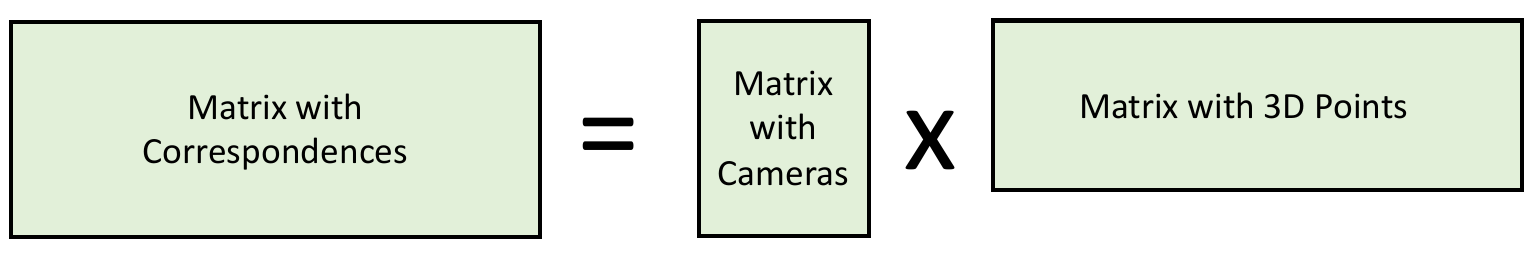}
    \caption{Projective factorization belongs to the first category in the proposed taxonomy, which considers structure and motion in a joint manner. 
    The idea is to collect all the correspondences in a single matrix, which is then factorized into a motion matrix and a structure matrix, under suitable assumptions.
    } 
    \label{fig:SAM_factorization}
\end{figure}

Let us collect all the (known) input correspondences, multiplied by the respective depths, into a single $3n \times p $ matrix $W$, which is called \emph{measurement matrix}:
\begin{equation}
    W =
\begin{bmatrix}
d_{11} \mathbf{m}_{11} & d_{12}  \mathbf{m}_{12}  & \dots & d_{1p}  \mathbf{m}_{1p}  \\
d_{21}  \mathbf{m}_{21} & d_{22}  \mathbf{m}_{22}  & \dots & d_{2p}  \mathbf{m}_{2p}  \\
\dots &  &  & \dots \\
d_{n1}  \mathbf{m}_{n1} & d_{n2}  \mathbf{m}_{n2}  & \dots & d_{np}  \mathbf{m}_{np} 
\end{bmatrix}
\label{eq_measurement_matrix_projective}
\end{equation}
where rows correspond to images and columns correspond to points.
In a similar way, we collect all the unknown cameras and unknown coordinates of 3D points into two matrices of sizes $3n \times 4$ and $4 \times p$, called the \emph{motion matrix} and the \emph{structure matrix}, respectively:
\begin{equation}
    P = \begin{bmatrix}
P_{1} \\
P_{2} \\
\dots \\
P_{n}
\end{bmatrix}, \quad
M = \begin{bmatrix}
M_{1} \
M_{2} \
\dots \
M_{p}
\end{bmatrix}.
\label{eq_motion_matrix}
\end{equation}
Note that Equation \eqref{eq_projection_depths}  holds for a single image point.
When considering all the points simultaneously, an equivalent compact form can be derived, which exploits the above notation:
\begin{equation}
    W = P S.
\end{equation}
In other terms, the measurement matrix is factorized into the product of two unknown matrices, that are the motion matrix and the structure matrix. This property is also visualized in Figure \ref{fig:SAM_factorization}.  In particular, this implies that $W$ has rank $4$ in general, which is low compared to its size.

In the presence of noise, however, the structure of $W$ is altered so that it does not exactly have rank 4. In this case, the best rank-4 approximation of $W$ is searched via the following optimization problem:
\begin{equation}
    \min_{P,S} || W - PS ||_F^2
\end{equation}
where $|| \ ||_F $ denotes the Frobenius norm. A closed-form solution can be obtained from the SVD of the measurement matrix, denoted by $W=U \Sigma V^{\mathsf{T}}$. Let $\widetilde{U}$ be the $3n \times 4$ sub-matrix obtained by keeping only the first four columns of $U$, let $\widetilde{V}^{\mathsf{T}}$ be the $4 \times p$ sub-matrix obtained by keeping only the first four rows of $V^{\mathsf{T}}$, and let $ \widetilde{\Sigma} $ be the $4 \times 4$ submatrix obtained by keeping only the fourth largest singular values. Then $ \widetilde{U} \widetilde{\Sigma} \widetilde{V}^{\mathsf{T}}$ represents the best (in the Frobenius norm sense) rank-4 approximation of $W$. Therefore the motion matrix is recovered as $P = \widetilde{U} \widetilde{\Sigma} $ while the structure matrix is recovered as $S=\widetilde{V}^{\mathsf{T}}$. Note that $ \widetilde{\Sigma}$ represents the global projective ambiguity, which could be placed in $S$ as well. Such estimates of motion and structure can be possibly refined via bundle adjustment, following the general scheme from Figure \ref{fig:SAM}.

To summarize, projective factorization recovers both structure and motion in a single step, under two main assumptions:
\begin{itemize}
    \item The projective depths are known;
    \item All points are visible in all images.
\end{itemize}
In practice, however, such assumptions are not satisfied.
In order to deal with unknown projective depths, a possible solution is by \emph{alternation}, as suggested in \cite{OliensisHartley07}:
\begin{itemize}
    \item With known projective depths, structure and motion can be recovered via SVD as explained above;
    \item With known structure and motion, recovering projective depths from Equation \eqref{eq_projection_depths} becomes a linear system.
\end{itemize}
The above steps are iterated until convergence or a maximum number of iterations is reached.
As for initialization, the projective depths can be set all equal to 1.

As concerns the case of missing correspondences, this translates into the fact that the measurement matrix is incomplete:
\begin{equation}
  W = \begin{bmatrix}
d_{11} \mathbf{m}_{11} & ?  & \dots & d_{1p} \mathbf{m}_{1p}  \\
d_{21} \mathbf{m}_{21} & d_{22} \mathbf{m}_{22}  & \dots & ?  \\
\dots &  &  & \dots \\
? & d_{n2} \mathbf{m}_{n2}  & \dots & d_{np} \mathbf{m}_{np} 
\end{bmatrix}.
\end{equation}
In order to deal with this challenging scenario, a possible approach is to estimate the missing entries of $W$ via rank minimization, therefore recovering a \emph{full} low-rank matrix to which the previous procedure can be applied \cite{MartinecPajdla05,JiaMartinez09,DaiLiHe13}. This task is also known as \emph{matrix completion} \cite{NguyenKimAl19}. 
Along this line, a more modern approach is proposed in \cite{MoranKoslowskyAl21,KathibKastenAl25}, where the authors combine ideas from the classical SfM formulation of projective factorization with deep learning approaches for matrix completion, using a loss that resembles the reprojection error. Special focus is given to respect the structure of the problem, in the sense that equivariance to permutations of cameras/points is guaranteed. 
A different permutation-equivariant architecture is proposed in \cite{BrynteIglesiasAl24} based on graph attention networks, using different types of features that resemble the structure of the measurement matrix -- see Equation \eqref{eq_measurement_matrix_projective}. Two regression heads are present at the end, one for the 3D points and another for the camera matrices. The main motivation behind these deep learning approaches \cite{MoranKoslowskyAl21,KathibKastenAl25,BrynteIglesiasAl24} is to learn SfM-specific primitives from several training examples, so that the model can be used for fast inference of the reconstruction for new and unseen sequences.

\begin{remark}
To summarize, the main advantages and disadvantages of projective factorization can be listed as follows. On one side, it is a global approach where all the cameras/points are considered simultaneously, thus promoting error compensation. Moreover, it works under weak assumptions, as the most general camera model (projective) is used. On the other side, projective factorization is memory demanding, requiring to store all the points at once.
\end{remark}

\subsection{Theoretical Conditions}
\label{subsec_SAM_theory}

We now discuss under which conditions the aforementioned SfM formulations are well posed in the sense of Remark \ref{remark_degeneracy}.

As concerns the \textbf{sequential approach}, well-posedness corresponds to situations where there are enough correspondences and degeneracies do not appear within the individual components, that are initialization by two-view reconstruction and model update by resection/intersection. 
Regarding two-view reconstruction, it is known that the fundamental matrix is not uniquely defined by image correspondences when the two cameras are related by pure rotation or all the 3D points and the two camera centres lie on a ruled quadric surface (a hyberboloid of one sheet, a cone, two planes, one plane) -- these scenarios are also referred to as \emph{critical configurations} \cite{HartleyZisserman04,Bratelund24b}. A similar result is available for two calibrated cameras \cite{KahlHartley02}.
As concerns triangulation, it is known that points in the baseline joining two cameras can not be reconstructed \cite{HartleyZisserman04}.
As concerns resection (i.e., camera recovery from 2D-3D correspondences), a {complete classification} of degeneracies is reported in \cite{HartleyZisserman04} for the case of an {uncalibrated} camera, comprising, e.g., the cases where the camera and points all lie on a twisted cubic. 
In the case of a {calibrated} camera, {all} degenerate configurations 
are listed in \cite{Thompson66}.

\smallskip

\begin{figure}[t]
    \centering  \includegraphics[width=1\linewidth]{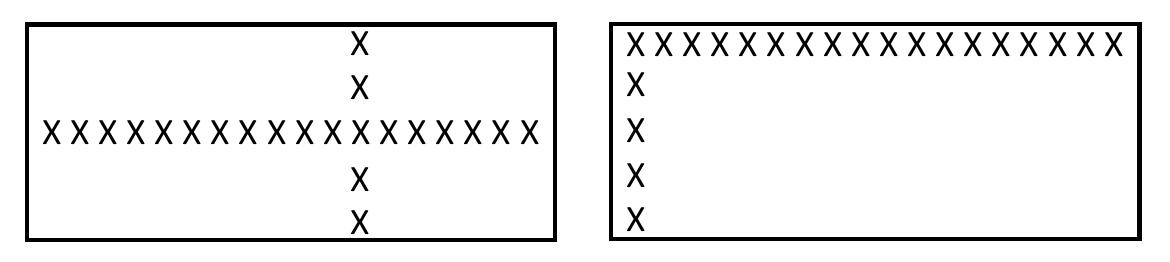}
    \caption{Examples of depth matrices that are cross-shaped. All entries are zeros except of those marked with $\times$. These matrices correspond to degenerate configurations in projective factorization.
    } 
    \label{fig:theory_depth}
\end{figure}

We now discuss degenerate cases for \textbf{projective factorization}, studied in 
\cite{NasihatkonHartleyAl15} under full visibility. 
With reference to the formulation with explicit depths -- the right part of Equation \eqref{eq_projection_depths} -- the authors of \cite{NasihatkonHartleyAl15} demonstrate that the solution to SfM is not always unique (up to a projective transformation) but multiple solutions (named \emph{false solutions}) may appear in some cases. 
Those degenerate situations correspond to annihilating a proper subset of the projective depths. This can be formalized by introducing the \emph{depth matrix}, that is the $n \times p$ matrix having the projetive depths $d_{ij}$ as entries. Its rows correspond to images and its columns correspond to 3D points.
Some specific forms of the depth matrix give rise to degeneracies:
\begin{itemize}
    \item the depth matrix has zero rows or zero columns;
    \item the depth matrix is cross-shaped (see Figure \ref{fig:theory_depth}).
\end{itemize}
Hence, in general, projective factorization is not always well-posed. 
However, it is well posed under additional assumptions:
it was proved in \cite{NasihatkonHartleyAl15} that, if the depth matrix is not in the above situations (i.e., it is not cross-shaped and it does not have zero rows/columns), then the SfM solution is unique, i.e., all configurations of cameras and 3D points yielding a common set of 2D image points are equal up to a projective transformation. This result is known as the \emph{Generalized Projective Reconstruction Theorem} \cite{NasihatkonHartleyAl15}. In fact, it is an extension of the classic \emph{Projective Reconstruction Theorem} \cite{HartleyZisserman04}, which states the same thesis under the assumption of non-zero projective depths. The authors of \cite{MagerandDel-Bue20} build upon \cite{NasihatkonHartleyAl15}, and develop a method for projective SfM which incorporates constraints to avoid the aforementioned degeneracies. The method is iterative and it resembles resection/intersection.

\section{Structure from Motion}
\label{sec_SFM}

\begin{figure}[t]
    \centering  \includegraphics[width=1\linewidth]{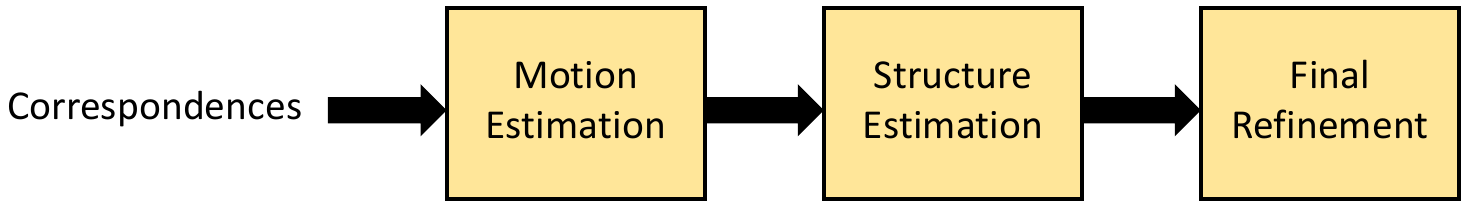}
    \caption{The second category in the proposed SfM taxonomy first recovers the motion and then the structure, starting from point correspondences. A final refinement (bundle adjustment) is applied at the end. The focus is on the motion step. Global SfM and its uncalibrated variation belong to this category.
    } 
    \label{fig:SFM}
\end{figure}

The second category in the proposed taxonomy (given in Figure \ref{fig:taxonomy}) separates structure and motion computation. 
Specifically,  \emph{first} the motion is recovered and \emph{then} the structure is computed, as sketched in Figure \ref{fig:SFM}.
Note that this type of approach is the only one that actually derives structure \emph{from} motion, for which the ``structure from motion'' term is truly appropriate.
The focus is on the motion step, motivated by the fact that recovering structure is a standard triangulation problem, once motion is known (see Table \ref{tab_resection}). The starting point is a set of correspondences and a bundle adjustment refinement is applied at the end, as usual.
The most popular approach within this category is Global SfM (Section \ref{subsec_globalSfM}), which operates with calibrated cameras. The uncalibrated case, instead, is less studied (Section \ref{subsec_multiviewF}). Theoretical conditions of these approaches are discussed in Section \ref{subsec_theory_SFM}.  In both cases, the problem is conveniently represented with a graph structure, as explained in Section \ref{subsec_viewing_graph}.

\subsection{The Viewing Graph}
\label{subsec_viewing_graph}

The \textbf{viewing graph} \cite{LeviWerman03} is a powerful tool to represent multiple cameras and their relationships, defined as follows:
\begin{itemize}
    \item each node represents a specific camera/image;
    \item an edge is present between two nodes if there are enough good correspondences between the two images, meaning that it is possible to reliably estimate the two-view geometry of the pair, encoded in the essential matrix (calibrated case) or fundamental matrix (uncalibrated case).
\end{itemize}
See Figure \ref{subfig:viewing_graph} for a visualization.
Several techniques are available to construct the viewing graph, such as
\cite{ShenZhuAl16,ShanChariAl18,CuiFragosoAl17,LiShiAl24}.
The viewing graph can be eventually optimized to remove edges that are likely to correspond to ambiguous or wrongly estimated two-view geometries. Specifically, in the calibrated case, wrong edges can be removed by loop closure detection, i.e., by exploiting the property that relative rotations (extracted from essential matrices) should compose to the identity over closed loops \cite{Govindu06,ZachKlopschitzAl10,BourmaudMegretAl14}. A more recent approach \cite{ManamGovindu24} jointly addresses false edge removal and graph sparsification -- the latter corresponds to the idea that reducing redundancy is good for the sake of efficiency without significant loss of reconstruction quality. That method can be applied both to calibrated and uncalibrated graphs, being based on an optimization problem involving the number of inlier correspondences for each edge.

We denote the viewing graph by $\mathcal{G}=(\mathcal{V}, \mathcal{E})$ where $\mathcal{V}$ denotes the vertex set and $\mathcal{E}$ denotes the edge set.
The number of nodes and edges are denoted by $n=|\mathcal{V}|$ and $e=|\mathcal{E}|$, respectively. A measure is associated to each edge $(i,j) \in \mathcal{E}$, that is the essential matrix $E_{ij}$ or the fundamental matrix $F_{ij}$, depending on the camera model. Note that the viewing graph can be thought of as an undirected graph: indeed, in practice, we typically have a measure for both edges $(i,j) \in \mathcal{E}$ and $(j,i) \in \mathcal{E}$, given by $F_{ji} = F_{ij}^{\mathsf{T}}$ and $E_{ji} = E_{ij}^{\mathsf{T}}$. 

Observe that, in this way, the challenging SfM task is divided into multiple (independent) subproblems that are easier to solve, as two-view geometries can be derived in closed form. In other terms, exploiting the viewing graph is tantamount to addressing SfM with a two-step approach:
\begin{enumerate}
    \item for each image pair, the essential/fundamental matrix is computed from image point correspondences;    
    \item all camera matrices are recovered starting from the two-view geometries estimated in Step 1, which are typically noisy in practice; outliers can also be present.
\end{enumerate}
Observe that in Step 1 we have local/relative information only, whereas in Step 2 we have global/absolute information.
The key idea in addressing Step 2 will be to exploit the consistency given by multiple two-view geometries in a principled way, as detailed in the sequel.
Note also that corresponding points are used in Step 1 only, but they are not considered in Step 2 (by most approaches), yielding a point-free formulation.

\subsection{The Calibrated Case - Global Approach}
\label{subsec_globalSfM}

Let us now consider the \textbf{calibrated case}, where the viewing graph formulation translates into the fact that we are given a redundant set of essential matrices (corresponding to the edges of the graph) and we have to recover the rotation matrices $R_i$ and translation vectors $\mathbf{t}_i$ of the cameras (corresponding to the nodes of the graph). Recall that the essential matrices can be decomposed into relative rotations and translation directions, representing the actual input of the motion step within the analyzed SfM pipeline. Before explaining how to recover cameras in practice, a convenient notation is introduced, which also gives insights on the estimation process. 

\begin{figure}[t]
    \centering  \includegraphics[width=1\linewidth]{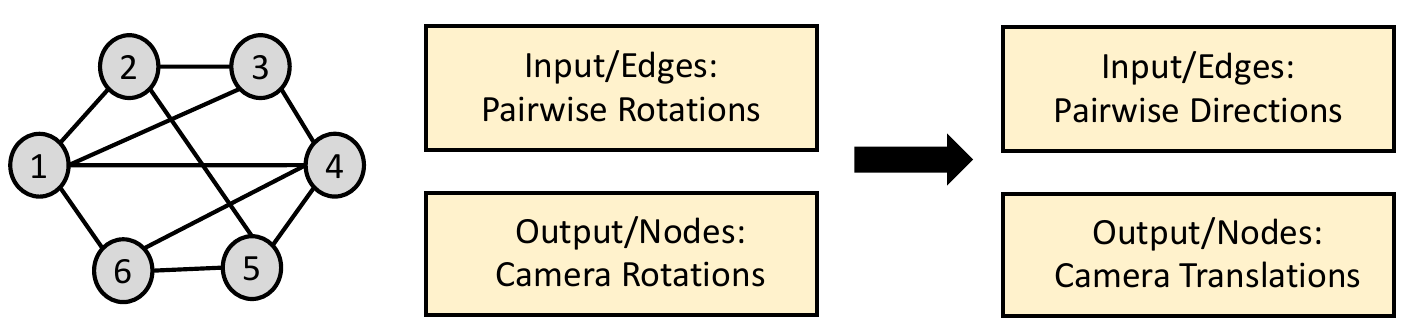}
    \caption{The second category in the proposed SfM taxonomy focuses on the motion step. In the calibrated case, two steps are typically employed: first, camera rotations are recovered from pairwise rotations; then, camera translations are recovered from pairwise directions. In both steps, input and output correspond to edges and nodes in the viewing graph representation, respectively. Recall that input quantities are extracted from essential matrices. 
    } 
    \label{fig:SFM_calibrated}
\end{figure}

Rotations and translations can be jointly encoded\footnote{The advantage of the SE(3) notation is that composition of transformations reduces to matrix multiplication.} by $4 \times 4$ matrices in the Special Euclidean Group, denoted by SE(3). Hence we introduce the following notation: 
\begin{equation}
    Z_{ij}=
\begin{bmatrix}
R_{ij} & \mathbf{t}_{ij} \\
\mathbf{0}^{\mathsf{T}} & 1
\end{bmatrix}, \
X_i=
\begin{bmatrix}
R_i & \mathbf{t}_i \\
\mathbf{0}^{\mathsf{T}} & 1
\end{bmatrix}, \
X_j=
\begin{bmatrix}
R_j & \mathbf{t}_j \\
\mathbf{0}^{\mathsf{T}} & 1
\end{bmatrix} 
\label{eq_SE3_motion}
\end{equation}
where $X_i$ represents the global transformation between a universal reference system and the reference system of camera $i$, while $Z_{ij}$ denotes the relative transformation bringing the reference frame of camera $j$ into that of camera $i$. Accordingly, the following compositional rule holds:
\begin{equation}
    Z_{ij} = X_i X_j^{-1}.
    \label{eq_composition_SE3}
\end{equation}
Equation \eqref{eq_composition_SE3} can be equivalently expressed as follows by separating rotation and translation components:
\begin{equation}
\begin{gathered}
R_{ij} = R_i R_j^{\mathsf{T}}  \\
\mathbf{t}_{ij} = -R_i R_j^{\mathsf{T}}  \mathbf{t}_j + \mathbf{t}_i
\end{gathered} 
\ \Longleftrightarrow \
\begin{gathered}
R_{ij} = R_i R_j^{\mathsf{T}}  \\
\underbrace{-R_i^{\mathsf{T}}  \mathbf{t}_{ij} }_{\mathbf{z}_{ij}}= \underbrace{- R_i^{\mathsf{T}}  \mathbf{t}_i }_{\mathbf{x}_i } + \underbrace{R_j^{\mathsf{T}}  \mathbf{t}_j}_{- \mathbf{x}_j} 
\end{gathered}
\label{eq_consistency_R_T}
\end{equation}
where $\mathbf{x}_i$ denotes the center of camera $i$ and $\mathbf{z}_{ij}$ denotes the relative displacement.
It is clear that rotations are independent of the translations, therefore they can be computed beforehand: most methods follow this paradigm, detailed below. See also Figure \ref{fig:SFM_calibrated} for a visualization. Alternative approaches will be discussed later in Remark \ref{remark_SE3}.

\smallskip 

As concerns \textbf{rotations}, the key idea is to exploit the rotation part of the consistency relation given in Equation~\eqref{eq_consistency_R_T}, namely: 
\begin{equation}
  R_{ij}=R_iR_j^{\mathsf{T}}
  \label{eq_consistency_R}
\end{equation}
which involves a specific edge $(i,j) \in \mathcal{E}$. Recall that the task is to recover camera rotations $R_1, \dots, R_n$ starting from multiple relative rotations $R_{ij}$ associated to the edges of the viewing graph. Before considering the general case of an incomplete graph, let us focus on the case of a complete graph, for simplicity of exposition.
Similarly to projective factorization, we store all the measures and the unknowns into matrices:
\begin{equation}
    Z = 
\begin{bmatrix}
I_3 & R_{12} & \dots & R_{1n} \\
R_{21} & I_3 & \dots & R_{2n} \\
\dots &  &  & \dots \\
R_{n1} & R_{n2} & \dots & I_3
\end{bmatrix}, \quad
X=
\begin{bmatrix}
R_{1} \\
R_{2} \\
\dots \\
R_{n}
\end{bmatrix}
\end{equation}
where $Z$ and $X$ have sizes $3n \times 3n
$ and $3n \times 3$ respectively. Hence Equation \eqref{eq_consistency_R} rewrites $Z=X X^{\mathsf{T}}$, which implies that $Z$ is symmetric, positive semi-definite, and it has low-rank (in fact, it has rank three). It was proved in \cite{Singer11} that the columns of $X$ are three eigenvectors of $Z$ with eigenvalue $n$, namely:
\begin{equation}
    Z X = n X.
    \label{eq_EIG_R_full}
\end{equation}
Hence, in the presence of noise, an approximate solution for $X$ is given by the three leading eigenvectors of $Z$. Such eigenvectors are then projected block-wise onto the space of rotations, denoted by SO(3), producing the camera rotations. This procedure is also known as the \emph{spectral method} \cite{Singer11}.

It can be proved that this approach generalizes to the case of an incomplete graph \cite{Arie-NachimsonKovalskyAl12}, where the $Z$ matrix is not fully specified but it has zero blocks in correspondence of missing edges. 
This is related to the following optimization problem:
\begin{equation}
\begin{gathered}
    \min_{R_1, \dots, R_n \in SO(3)} \sum_{(i,j) \in \mathcal{E}} || R_{ij} - R_i R_j^{\mathsf{T}} ||_F^2 \\ 
    \Longleftrightarrow
    \max_{X \in SO(3)^n} \text{trace}(X^{\mathsf{T}}Z X)
    \end{gathered}
    \label{eq_R_objective}
\end{equation}
which minimizes the discrepancy between the left and right sides in Equation \eqref{eq_consistency_R}.
Specifically, the spectral method corresponds to a \emph{relaxation} of the above objective, i.e., some constraints are ignored and a simpler problem is considered: if we force the optimization variable to satisfy $X^{\mathsf{T}}X=I_3$, then  a standard Rayleigh problem is obtained, whose solution is given by the three leading eigenvectors of $Z$. A similar approach is adopted in \cite{MartinecPajdla07} based on a linear system. Robust extensions of the spectral method are also available \cite{ArrigoniRossiAl16}, where a weighted graph is considered in a combination with Iteratively Reweighted Least Squares (IRLS) \cite{HollandWelsch77}.

With reference to Problem \eqref{eq_R_objective}, other relaxations were considered in the literature. One example is the \emph{rank relaxation}, which  resembles the idea of low-rank matrix completion also employed by projective factorization, both in the classical sense \cite{ArrigoniRossiAl18} and exploiting the implicit regularization given by deep matrix factorization  \cite{TejusZaraAl23}. 
Another example is the \emph{semi-definite relaxation}, where the $Z$ matrix is enforced to be positive semi-definite \cite{Singer11, Arie-NachimsonKovalskyAl12} ignoring the other constraints. A more general objective can also be considered \cite{HartleyTrumpfAl13}:
\begin{equation}
        \min_{R_1, \dots, R_n \in SO(3)} \sum_{(i,j) \in \mathcal{E}} d( R_{ij} , R_i R_j^{\mathsf{T}})^k
        \label{eq_rotation_synch}
\end{equation}
where typically $k=1$ or $k=2$, and $d$ denotes a distance between rotations (e.g., the angular distance, the Frobenius norm, the quaternion distance, etc.). Different parameterizations can be chosen besides rotation matrices, such as unit quaternions \cite{Govindu01} or the angle-axis representation.
Problem \eqref{eq_rotation_synch} is also called \emph{rotation averaging} or \emph{rotation synchronization}.

Approaches to solve \eqref{eq_rotation_synch} include the Levenberg-Marquardt algorithm \cite{CrandallOwensAl11} or Lie-group optimization \cite{ChatterjeeGovindu13,ShiLerman20}.
More recent methods focus on global optimality  \cite{ErikssonOlssonAl18,DellaertRosenAl20,ParraChngAl21,MoreiraMarquesAl21}, including uncertainty into the formulation \cite{BirdalArbelAl20,ZhangLarssonAl23}, exploiting additional information like the gravity direction \cite{PanPollefeysAl24} or  initialization techniques for nonlinear minimization \cite{LeeCivera22}. Other authors explored a deep learning framework in a supervised manner with graph neural networks \cite{PurkaitChinAl20,YangLiAl21,LiLing21,LiCuiAl22}.
A different paradigm is adopted in \cite{HartleyAftabAl11} where an iterative method is developed that updates each camera rotation in turn based on its neighboring edges. See also the survey \cite{TronZhouAl16} for additional references.

\begin{remark}
Observe that optimizing over rotations is a hard task: as observed in \cite{HartleyTrumpfAl13},  Problem \eqref{eq_rotation_synch} is non convex and may have multiple local minima in separate basins of attraction.
Further theoretical analysis is given in \cite{BoumalSingerAl14,WilsonBindelAl16,WilsonBindel20}. In particular, the authors of \cite{WilsonBindelAl16} define a concrete measure quantifying how hard a rotation averaging problem is:
\begin{equation}
    \frac{\lambda_2(L)}{n}
\end{equation}
where $n$ is the number of cameras and $\lambda_2(L)$ denotes the second-smallest eigenvalue of the Laplacian matrix associated to the viewing graph. The latter is defined as $L=D-A$, where $D$ and $A$ denote the adjacency matrix and the degree matrix of the graph, respectively.
Based on such a measure, it is observed in \cite{WilsonBindelAl16} that smaller and well-connected problems are easier, but larger/noisier problems are typically difficult, as expected. %
These theoretical findings motivate the development of  partitioned approaches \cite{BhowmickPatraAl14,DalcinMagriAl21,ZhuZhangAl18}.
\end{remark}

\smallskip

We now consider camera \textbf{translations} and focus on the translation part of Equation \eqref{eq_consistency_R_T}, namely: 
\begin{equation}
    \mathbf{z}_{ij} = \mathbf{x}_i - \mathbf{x}_j
\label{eq_consistency_C}
\end{equation}
where $\mathbf{x}_i$ denotes the unknown center of camera $i$ (also known as \emph{location} or \emph{position}) and $\mathbf{z}_{ij}$ represents the relative displacements between the two cameras in edge $(i,j) \in \mathcal{E}$. Recall that the recovery of camera centers (with known rotations) implies the knowledge of camera translations, see Equation~\eqref{eq_centre}. A key observation is that we can not compute centers by just collecting equations like \eqref{eq_consistency_C} for all the edges in the viewing graph and solving the resulting linear system, since relative displacements are not fully known:
\begin{equation}
    || \mathbf{t}_{ij} || = || \mathbf{z}_{ij} || = ? 
\end{equation}
Indeed, as already observed, the magnitude of relative translations are unknown but only the directions can be extracted from essential matrices. In other terms, the goal here is to recover the position of the cameras (associated to nodes in the viewing graph), where pairs of cameras can measure the direction of the line joining their locations (corresponding to the edges). For edge $(i,j) \in \mathcal{E}$ such direction is given by:
\begin{equation}
    \mathbf{u}_{ij} = \frac{\mathbf{z}_{ij} }{ || \mathbf{z}_{ij} || } = \frac{\mathbf{x}_{i}-\mathbf{x}_{j} }{ ||\mathbf{x}_{i}-\mathbf{x}_{j} || }  .
    \label{eq_direction}
\end{equation}
The geometric meaning of the above constraint is that  $ \mathbf{u}_{ij} $ and $ (\mathbf{x}_{i}-\mathbf{x}_{j}) $ are parallel vectors. This can be rewritten as:
\begin{equation}
 \mathbf{u}_{ij} \times (\mathbf{x}_{i}-\mathbf{x}_{j} ) = 0   \Longleftrightarrow 
    [ \mathbf{u}_{ij}]_{\times} (\mathbf{x}_{i}-\mathbf{x}_{j} ) = 0. 
    \label{eq_bearing_skew}
\end{equation}
By collecting equations like \eqref{eq_bearing_skew} from all the edges, a homogeneous linear system of the following form is obtained:
\begin{equation}
    S \mathbf{x} = \mathbf{0}
    \label{eq_centre_all}
\end{equation}
where $S$ has size $3e \times 3n$ and $\mathbf{x}$ is an unknown vector of length $3n$ given by the vertical concatenation of all the $\mathbf{x}_i$.
Recall that $n$ is the number of nodes and $e$ is the number of edges.
Hence, in the presence of noise, an approximate solution for the camera centers is found 
via least-squares \cite{Govindu01,BrandAntoneAl04}.

\begin{remark} 
It was noted in \cite{OzyesilSingerAl15,OzyesilSinger15} that the least-squares solution suffers from the ``clustering effect'', meaning that camera centers can collapse to a few (wrong) locations. A possibility to overcome this drawback is to include additional constraints in the global objective, such as: 
\begin{equation}
\begin{gathered}
    \min_{\mathbf{x}_{i}, \dots, \alpha_{ij}} \sum_{(i,j) \in \mathcal{E}} ||\mathbf{x}_{i}-\mathbf{x}_{j}- \alpha_{ij} \mathbf{u}_{ij}|| \\
\text{such that } \sum_{i=1}^n \mathbf{x}_i=\mathbf{0}, \quad \alpha_{ij} \ge 1 .
\end{gathered}
\label{eq_clustering}
\end{equation}
Note that the first constraint fixes the global translation ambiguity while the second constraint forces distances between camera centers to be sufficiently large. Problem \eqref{eq_clustering} can be solved with semi-definite programming \cite{OzyesilSingerAl15} or quadratic programming \cite{OzyesilSinger15}.
Further theoretical studies are reported in \cite{ManamGovindu23,HeRuanAl25}. 
As an example, the authors of \cite{ManamGovindu23} observe that small angles (between two pairwise translation directions) may cause numerical instabilities: this is encapsulated by the condition number of a proper matrix containing all angles. Additional stability analysis is given in \cite{HeRuanAl25}. 
\end{remark}

Alternative ways to approach camera location recovery include a bilinear formulation \cite{ZhuangCheongAl18} or triplet-based formulations \cite{JiangCuiAl13,MoulonMonasseAl13}. Popular optimization schemes adopted in the literature comprise Riemannian gradient descent \cite{TronVidal14}, the Levenberg-Marquardt algorithm \cite{CrandallOwensAl11,WilsonSnavely14}, the alternating direction method of multipliers (ADMM) \cite{GoldsteinHandAl16}, and adaptive annealing \cite{SidharthaGovindu24}. 
There is also notable research on de-noising pairwise directions \cite{ManamGovindu22} or detecting/removing highly corrupted directions \cite{WilsonSnavely14,ShiLerman18}, in order to improve the overall accuracy of the estimation process.

\begin{remark}
The methods discussed so far  split the motion recovery step into rotation and translation, following Equation~\eqref{eq_consistency_R_T}. Alternatively, rotations and translations can be jointly recovered in $SE(3)$. This idea, although theoretically relevant, implies solving for the unknown scales of pairwise directions. A possible approach is by alternation \cite{Govindu06}, where all the scales are initialized to 1 and iteratively updated based on the current estimate of camera motion, similarly to depth recovery in projective factorization \cite{OliensisHartley07}. Another approach consists in explicitly recovering the scales in one shot via cycle consistency \cite{ArrigoniFusielloAl15}, prior to motion estimation. %
An alternative to the $SE(3)$ representation consists in working directly with essential matrices, as done in \cite{KastenGeifmanAl19b}: this approach resembles the one with uncalibrated cameras and fundamental matrices detailed in Section \ref{subsec_multiviewF}, hence it is not explained here.
\label{remark_SE3}
\end{remark}

\begin{remark}
The pipeline described in this section is also known as \textbf{Global Structure from Motion} since the whole viewing graph is considered\footnote{This, however, does not imply that only global operations are performed, but intermediate \emph{local} steps might be involved (see, e.g., \cite{HartleyAftabAl11}).}. Hence, provided that there is enough redundancy in terms of amount of edges, a fair distribution of the errors among the cameras is typically achieved, as opposed to the sequential approach which is subject to error propagation.
Recall that, after computing two-view geometries and storing them as edges in the viewing graph, points are not used for motion estimation: on one side, this implies that accuracy is suboptimal compared to methods using points (such as those discussed in Section \ref{sec_SAM}); on the other side, the absence of points implies that global SfM approaches are usually faster and less memory demanding than sequential/hierarchical ones.
Overall, global SfM methods can be seen as an efficient and effective way of computing approximate camera motion to be subsequently refined by bundle adjustment.
\label{remark_global_sfm}
\end{remark}

\begin{remark}
Although most existing methods follow a ``point-free'' paradigm, as also recalled in the previous remark, it is important to observe that -- in principle -- the viewing graph can be generalized to a larger graph which contains both locations of cameras and 3D points as nodes, and both camera-camera and 3D point-camera relationships as edges. Indeed, directional information about some of the vectors connecting cameras to image points is available. This is observed (for example) in \cite{BrandAntoneAl04,WilsonSnavely14,TaoCuiAl24}. Hence the methods discussed above can be (in principle) adapted to the joint recovery of 3D points and camera locations, avoiding the need for a subsequent triangulation. A recent approach  implementing this scheme is GLOMAP \cite{ PanBarathAl24}, which achieves comparable results to COLMAP \cite{SchonbergerFrahm16}, thereby constituting a valid alternative to sequential SfM in practice.    
\end{remark}

\begin{remark}
Finally, it is worth mentioning a recent approach based on deep learning, named VGGSfM \cite{WangKaraevAl24}. Note that is does not comply with the assumptions considered in this survey, as it directly estimates point correspondences, whereas classical methods typically assume that correspondences have been pre-computed (see Section \ref{sec_problem}). However, VGGSfM resembles the paradigm from Figure~\ref{fig:SFM}: first, the motion is recovered (via a Transformer) and then the structure is computed (via a differentiable triangulation). Computing correspondences (preliminary step) and a bundle adjustment refinement (final step) are all integrated into a fully differentiable pipeline, which is trained in a supervised manner.
\label{remark_VGGSFM}
\end{remark}

\begin{figure}[t]
    \centering  \includegraphics[width=0.7\linewidth]{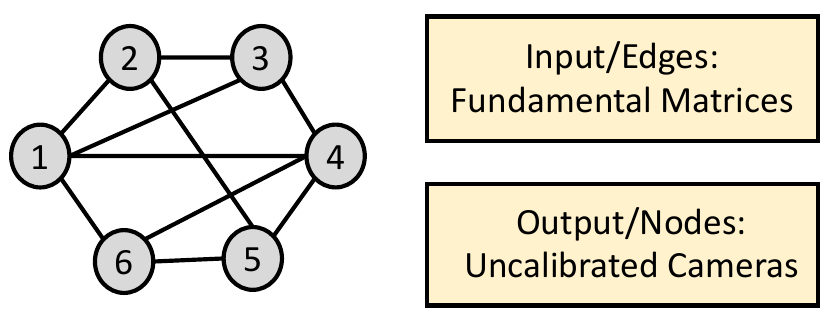}
    \caption{The second category in the proposed SfM taxonomy focuses on the motion step. In the uncalibrated case, this translates into estimating cameras starting from fundamental matrices. With reference to the viewing graph representation, the input is associated with edges, whereas the output is associated with nodes, as in the calibrated case.
    } 
    \label{fig:SFM_uncalibrated}
\end{figure}

\subsection{The Uncalibrated Case - Multi-view Fundamental Matrix}
\label{subsec_multiviewF}

Let us now consider the general pipeline from Figure~\ref{fig:SFM} in the \textbf{uncalibrated case}. Recall that the focus is on the motion step and the problem can be represented as a viewing graph, as explained in Section \ref{subsec_viewing_graph}. In this scenario, the input is a redundant/noisy set of fundamental matrices (corresponding to the edges of the graph) and the objective is to recover the camera matrices (corresponding to the nodes of the graph). See Figure \ref{fig:SFM_uncalibrated} for a visualization.

Similarly to the rotation step in the calibrated case, we store all the measures in a $3n \times 3n$ block-matrix, which resembles the structure of the adjacency matrix of the viewing graph:
\begin{equation}
    F = 
\begin{bmatrix}
0 & F_{12} & \dots & F_{1n} \\
F_{21} & 0 & \dots & F_{2n} \\
\dots &  &  & \dots \\
F_{n1} & F_{n2} & \dots & 0
\end{bmatrix}.
\end{equation}
We assume here that the graph is complete and discuss the case of missing data at a later point. $F$ is also known as the \emph{multi-view fundamental matrix} \cite{SenguptaAmirAl17,KastenGeifmanAl19} as it encapsulates information coming from all the individual fundamental matrices between pairs of images. Note that $F$ is symmetric. Using this notation, the SfM objective becomes recovering the motion matrix $P$ -- see Equation \eqref{eq_motion_matrix} -- starting from $F$.

Observe that the relative measures and the unknown variables (i.e., the fundamental matrices and the camera matrices) have a different form (and even a different size), which makes the problem challenging. In the case of rotations, instead, both the input and the desired output were rotation matrices, allowing the usage of group operations, at the basis of several algorithms (such as \cite{ChatterjeeGovindu13,Singer11}). Notwithstanding this, it is still feasible to derive a \emph{spectral solution}, under suitable assumptions.
Specifically, it is proved in \cite{KastenGeifmanAl19} that the multi-view fundamental matrix is \emph{consistent} (i.e., there exists a set of cameras producing the input fundamental matrix) if and only if the following conditions hold:
\begin{equation}
    \begin{gathered}
        \text{rank}(F)=6 \\
        F \text{ has exactly 3 positive and 3 negative eigenvalues} \\
\text{rank}(F_i)=3, \  
\text{where } F_i \text{ is the i-th block row.}
    \end{gathered}
\label{eq_consistency_F}
\end{equation}
It was later shown that the second condition is redundant and can be dropped \cite{BratelundRydell23}.
The third condition means that the camera centers are not all collinear. 
If \eqref{eq_consistency_F} holds, then camera matrices can be derived in closed-form via a proper manipulation of the six eigenvectors of $F$ corresponding to nonzero eigenvalues, as detailed in \cite{KastenGeifmanAl19}. 

Note that the characterization from \eqref{eq_consistency_F} holds when each $F_{ij}$ is properly scaled and the graph is complete, severely limiting the usage of the spectral method in practice. A possible approach to recover the unknown scale associated with each fundamental matrix is discussed in \cite{SenguptaAmirAl17}, where a suitable objective is defined in order to refine the input fundamental matrices while at the same time computing the unknown scales, under the constraint that the multi-view $F$ is consistent. This objective is minimized with ADMM, which, however, is sensitive to initialization, as observed by the authors. 

A different approach is employed in \cite{KastenGeifmanAl19} where the authors observe that -- \emph{when restricted to three views only} -- the multi-view fundamental matrix is scale-invariant, thus motivating a partitioned (triplet-based) approach that can inherently cope both with an incomplete graph, without needing to explicitly compute the unknown scales. The method follows these steps:
\begin{enumerate}
    \item A \emph{triplet cover} of the viewing graph is extracted, namely a subset of all possible triplets (i.e., loops of length three) with the property that each triplet shares (at least) an edge with another one;
    
    \item The input fundamental matrices are \emph{refined} by solving (with ADMM) a rank-constrained optimization problem\footnote{
    Note that -- in theory -- consistency over triplets implies global consistency (see Theorem 2 in \cite{KastenGeifmanAl19}). %
    However, since soft constraints are employed by ADMM, there are no guarantees that the refined fundamental matrices will be actually consistent. However, they are closed to being consistent in practice.} defined over triplets, based on \eqref{eq_consistency_F};
    
    \item For each triplet, the three camera matrices are recovered from the refined fundamental matrices, using the {spectral method} detailed above; this results into multiple \emph{independent} three-view reconstructions;

    \item Camera matrices of individual triplets are brought into a common projective frame, by sequentially concatenating $4 \times 4$ projective transformations over consecutive triplets.    
\end{enumerate}
Step 4 has been recently improved by \cite{MadhavanFusielloAl24} by replacing the sequential propagation with a global approach. Other  attempts to recover cameras from fundamental matrices include \cite{SinhaPollefeysAl04,ColomboFanfani21}: these methods first recover cameras in a triplet with closed-form expressions, and then provide induction steps to manage larger viewing graphs made by multiple triplets.

\begin{remark}
The approach detailed in this section can be viewed as the uncalibrated version of Global SfM, hence it is also referred to as \textbf{Global Projective Structure from Motion}. Accordingly, the advantages/disadvantages of this scheme with respect to other categories (like sequential SfM or projective factorization) are similar to the calibrated case (see Remark~\ref{remark_global_sfm}).
However, the uncalibrated scenario is, in general, much more difficult than the calibrated one, as weaker assumptions are made, in line with the discussion in Section \ref{subsec_theory_SFM}.
\end{remark}

\begin{figure}[t]
    \centering  \includegraphics[width=1\linewidth]{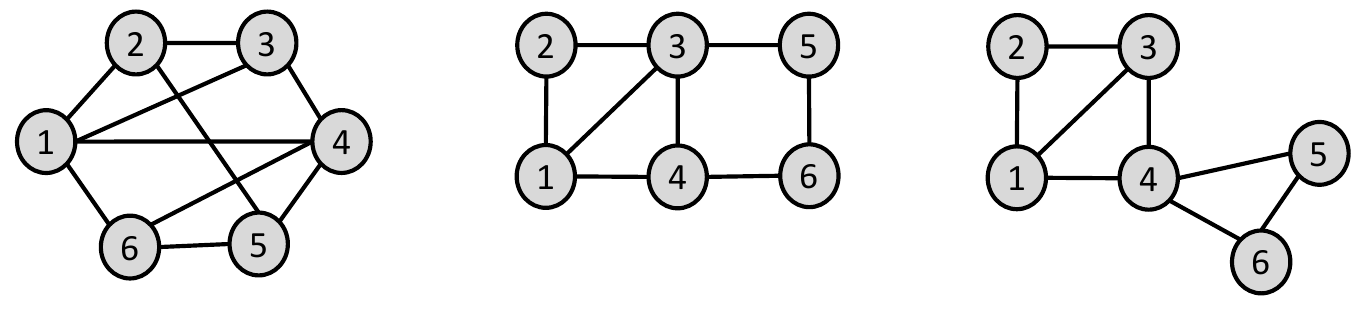}
    \caption{Examples of viewing graphs with six nodes. The graph on the left is non degenerate for both calibrated and uncalibrated scenarios (i.e., solvable and parallel rigid), meaning that essential/fundamental matrices (on the edges) uniquely determine the cameras (on the nodes). The graph on the middle is degenerate for uncalibrated cameras but not for calibrated ones. %
    The graph on the right is degenerate for both calibrated and uncalibrated cameras, due to the presence of an articulation point (node number 4).
    } 
    \label{fig:theory_SFM}
\end{figure}

\subsection{Theoretical Conditions}
\label{subsec_theory_SFM}

We now discuss degenerate cases for the second category of SfM methods in the proposed taxonomy (Figure \ref{fig:SFM}), comprising calibrated/uncalibrated global SfM. The importance of understanding degeneracies was highlighted in Remark \ref{remark_degeneracy}.
In this scenario, degenerate configurations can be formalized in terms of the viewing graph and they differ between calibrated and uncalibrated cameras, as detailed in the sequel. 
A common assumption is the \emph{genericity assumption}, meaning that cameras are assumed to be generic. For example, random cameras are generic whereas cameras lying on a plane or on a straight line are not.
Under this assumption,  degeneracies solely depend on the {graph} and not on the actual fundamental/essential matrices.

\smallskip

In the \textbf{calibrated case}, a graph is non-degenerate if and only if the available two-view relationships uniquely (up to a single transformation) determine the cameras, i.e., there is a unique solution to SfM under the global formulation. Otherwise, the graph is called degenerate, meaning that there are multiple solutions. Existing work address this important theoretical question by separating the problem into rotation and translation, similarly to most practical approaches.
Due to the genericity assumption, non-degenerate graphs can be identified by studying in which cases cameras are uniquely determined by the constraints in a noiseless scenario. 
As concerns {rotations}, it is known that a unique solution to Equation \eqref{eq_consistency_R} is possible under the assumption of a \textbf{connected} graph \cite{HartleyTrumpfAl13}.
In a noiseless scenario, such unique solution can be easily computed by propagating the consistency constraint -- rewritten as $ R_i = R_{ij} R_j $ -- along a spanning tree of the viewing graph. If a graph is not connected, then extracting the largest connected subgraph is a standard procedure \cite{Harary72}.

As concerns translations, camera locations can be uniquely recovered from pairwise directions, i.e., there is a unique solution to Equation \eqref{eq_bearing_skew}, if and only if: 
\begin{equation}
    \text{rank} (S) = 3n -4
\label{eq_rigidity_rank}
\end{equation}
where $n$ is the number of nodes in the graph and $S$ is defined in Equation \eqref{eq_centre_all}. Observe that the rank is expected to drop by four due to the scale and translation ambiguity, hence degenerate graphs\footnote{Alternatively, one could fix the translation/scale ambiguity (e.g., by adding extra equations) so that non-degenerate graphs correspond to full rank.} correspond to cases where $   \text{rank} (S) < 3n -4$. 
In addition to the description from Equation \eqref{eq_rigidity_rank}, a combinatorial characterization of non-degenerate graphs is also available: it was observed in \cite{OzyesilSinger15} that non-degenerate graphs correspond to \textbf{parallel rigid} graphs. This finding is based on the fact that recovering camera centers from pairwise directions is a particular instance of a well-studied problem named \emph{bearing-based localization} \cite{ZhaoZelazo16}. Being parallel rigid means that all configurations (i.e., assignment of locations to the nodes) that have parallel edge directions are related by translation and/or scale. See \cite{ArrigoniFusiello18} for a survey of parallel rigidity and see Figure~\ref{fig:theory_SFM} for some visualizations.

In order to test parallel rigidity in practice, Equation \eqref{eq_rigidity_rank} is typically employed following these steps:
\begin{enumerate}
    \item A set of camera centers $\mathbf{c}_1, \dots, \mathbf{c}_n$ is sampled at random;
    \item The centers from Step 1 are used to derive pairwise directions for the edges using Equation \eqref{eq_direction}; 
    \item The linear system given in Equation \eqref{eq_centre_all} is built, treating the directions from Step 2 as known values;
    \item If Equation \eqref{eq_rigidity_rank} holds than the viewing graph is parallel rigid, otherwise it is degenerate (also called \emph{flexible)}.
\end{enumerate}
Note that sampling cameras at random is a practical way to comply with the genericity assumption. The above procedure means that, given a graph and a random configuration of cameras, we check if those cameras are the only solution with those directions, or there are more solutions. If a graph is degenerate, then it is possible to partition the edges into maximal components that are parallel rigid 
\cite{TronCarloneAl15,KennedyDaniilidisAl12}. The approach from \cite{KennedyDaniilidisAl12} group edges by clustering rows in the null-space of $S$, that is bigger than expected in a degenerate graph. Necessary conditions for parallel rigidity are also available: for example, a parallel rigid graph must have at least $(3n-4)/2$ edges; this serves as an auxiliary (straightforward) tool to discard degenerate graphs.
As shown in \cite{ArrigoniFusiello18}, the most common sources for degeneracies in real SfM graphs are the presence of pendent edges and articulation points. An example of the latter can be appreciated in Figure~\ref{fig:theory_SFM} (right).

\smallskip 

In the \textbf{uncalibrated case}, a viewing graph is non-degenerate if and only if the available fundamental matrices uniquely (up to a single projective transformation) determine the cameras. 
In this case, the graph is called \emph{solvable} \cite{TragerHebertAl15} or \emph{solving} \cite{LeviWerman03}. Otherwise, the graph is called degenerate or non-solvable, meaning that there are multiple solutions\footnote{Here, the multiplicity of solutions can be either finitely many or infinitely many, since we are concerned with polynomial equations. With calibrated cameras, instead, the problem admits a linear formulation, meaning that the number of solutions is either unique or infinitely many, but not finitely many.} to projective SfM under the global formulation. 
Due to the genericity assumption, non-degenerate graphs can be identified by studying in which cases cameras are uniquely determined by the constraints in a noiseless scenario, as for the calibrated case. See Figure \ref{fig:theory_SFM} for examples of solvable and non solvable graphs.

Specifically, a possible approach is to exploit Result 9.12 in \cite{HartleyZisserman04}, stating that a non-zero matrix $F_{ij}$ is the fundamental matrix corresponding to a pair of cameras $P_i$ and $P_j$ if and only if the matrix $   P_j^{\mathsf{T}} F_{ij} P_i  $ is skew-symmetric. This condition can be rewritten as:
\begin{equation}
   P_j^{\mathsf{T}}   F_{ij} P_i  + P_i^{\mathsf{T}}   F_{ij}^{\mathsf{T}}  P_j = 0.
\label{eq_consistency_F_P_HZ}
\end{equation}
Hence, checking solvability reduces to computing the number of solutions of the \emph{polynomial} system obtained by collecting equations like \eqref{eq_consistency_F_P_HZ} for all the edges in the viewing graph, where fundamental matrices are known and cameras are unknown \cite{ArrigoniFusielloAl24}. In practice, thanks to the genericity assumption, those fundamental matrices are derived from random cameras, similarly to the procedure for the calibrated case, hence the question is how many camera configurations are compatible with those (noiseless) fundamental matrices. Extra equations can be possibly added to fix the projective and scale ambiguity, so that a solvable graph corresponds to a unique solution: this can be checked with symbolic solvers like Grobner basis.
Other polynomial formulations are available in the literature, which explicitly reason in terms of the problem ambiguities, modeled as $4 \times 4$ projective matrices \cite{TragerOssermanAl18,ArrigoniFusielloAl21}.

However, solving polynomial equations is extremely hard, since the complexity is exponential in the number of variables in the worst case \cite{Dube90}. To overcome this challenge, some authors focus on an approximation of solvability named \emph{finite solvability}, corresponding to a finite number of solutions (instead of uniqueness). This can be checked very efficiently, as it reduces to computing the rank of a proper matrix \cite{TragerOssermanAl18,ArrigoniPajdlaAl23,ArrigoniFusielloAl24}. However, it is not possible to know the exact number of solutions in this way, i.e., it is not possible to distinguish between one solution and  (e.g.) two distinct solutions.
Notwithstanding this, finite solvability represents a good proxy for solvability in practice, allowing to handle large-scale datasets. It is also possible to partition an unsolvable graph into maximal components that are finitely solvable, by extending the procedure for calibrated cameras developed in \cite{KennedyDaniilidisAl12}. See \cite{ArrigoniPajdlaAl23,ArrigoniFusielloAl24} for more details.

Useful tools to complement the approximation induced by finite solvability are necessary conditions or sufficient conditions for solvability. The former permit to easily discard degenerate graphs, whereas the latter permit to prove that certain graphs are solvable, although not covering all possible cases.
For example, it was proved in \cite{TragerHebertAl15} that graphs built from a triangle by adding vertices of degree two one at a time are solvable. This implies, in particular, that the triplet-based graphs used by \cite{SinhaPollefeysAl04,ColomboFanfani21,KastenGeifmanAl19,MadhavanFusielloAl24} are solvable (see Section \ref{subsec_multiviewF} for an explanation of those techniques). However, note that solvable graphs may also not contain triplets at all (see \cite{LeviWerman03} for some examples). Additional constructive approaches are available \cite{RudiPizzoliAl11,TragerOssermanAl18}. 
Concerning necessary conditions, it was proved that a solvable graph satisfies these conditions: 
\begin{itemize}
    \item it has at least $(11n-15)/7$ edges \cite{TragerOssermanAl18};
    \item it is biconnected \cite{TragerOssermanAl18};
    \item all the vertices have degree at least two and no two adjacent vertices have degree two (if $n > 3$) \cite{LeviWerman03}. 
\end{itemize}
Overall, identifying degenerate graphs in uncalibrated scenarios is much more difficult than the calibrated case.

\smallskip 

As concerns the connection between calibrated and uncalibrated viewing graphs, it was proved in \cite{ArrigoniFusielloAl22} that: if a graph is solvable (i.e., non degenerate with uncalibrated cameras) then it is also parallel rigid (i.e., non degenerate with calibrated cameras), as expected. However, the converse does not hold, namely there exist graphs that are parallel rigid but non solvable: see Figure~\ref{fig:theory_SFM} (middle) for an example.

\section{Structure without Motion}
\label{sec_SWM}

\begin{figure}[t]
    \centering  \includegraphics[width=1\linewidth]{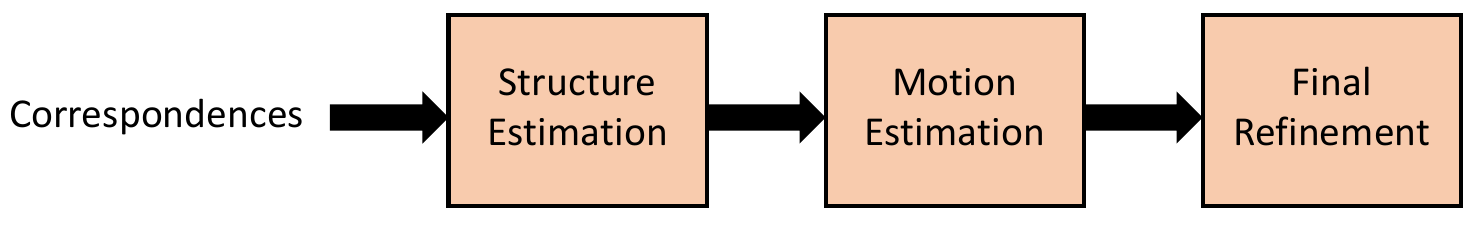}
    \caption{The third category in the proposed SfM taxonomy directly estimates the structure from point correspondences. Motion estimation is optional and can be performed subsequently. A final refinement (bundle adjustment) is applied at the end. Very few approaches belong to this category.
    } 
    \label{fig:SWM}
\end{figure}

The third (and last) category in the proposed taxonomy directly estimates the structure starting from point correspondences, \emph{without} recovering camera motion, as shown in Figure~\ref{fig:SWM}. If needed, motion can be estimated \emph{after} the structure as a byproduct: indeed, the scene structure translates into a set of 2D-3D correspondences, hence motion follows from the solution of a standard resection problem (see Table~\ref{tab_resection}). A bundle adjustment refinement is applied at the end, as usual.
This kind of approach is less studied than the other categories in the proposed taxonomy (see Figure \ref{fig:taxonomy}), and it counts a limited amount of works, focused on the calibrated case.

\smallskip

\begin{figure}[t]
    \centering  \includegraphics[width=1\linewidth]{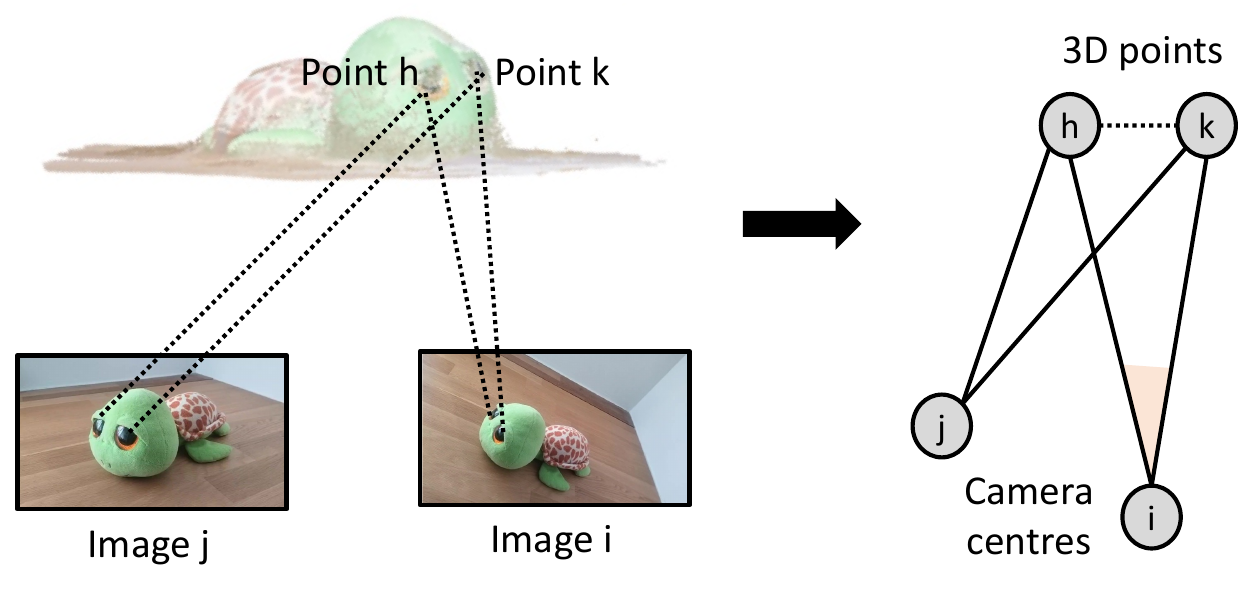}
    \caption{The third category in the proposed SfM taxonomy focuses on structure estimation. A convenient graph representation is used, where nodes represent either 3D points or camera centers, and edges encode different types of relationships, like camera-point or point-point. The basic equation used by \cite{Li10} considers a triangle made by two 3D points and one camera.
    } 
    \label{fig:SWM_graph}
\end{figure}

To the best of our knowledge, the first method following this paradigm is \cite{Li10}, which is reviewed here. The theoretical conditions of this approach will be discussed later in Remark~\ref{remark_theory_SWM}. The key idea of \cite{Li10} is to put the focus on computing the unknown \emph{distances} between 3D points. The rationale behind this idea is that -- with known distances -- the 3D coordinates can be recovered by standard techniques, such as Multi-Dimensional Scaling (MDS) \cite{AspnesErenAl06}. To this aim, a graph representation of the problem is employed:
\begin{itemize}
    \item Nodes are either 3D points or camera centres;
    \item Two 3D points are connected by an edge if their imaged points are simultaneously seen by two (or more) cameras. In this case, the corresponding edges between the camera centres and the point pair are also drawn (Figure \ref{fig:SWM_graph}). 
\end{itemize}
The basic equation involves a triplet in this graph made by two 3D points (indexed by $h$ and $k$) and one camera (indexed by $i$), which is viewed as a triangle. The \emph{law of cosines} applies:
\begin{equation}
     l_{hi}^2 + l_{ki}^2 - 2 l_{hi} l_{ki} \cos(\theta_{hik}) = \ell_{hk}^2
     \label{eq_law_cosines}
\end{equation}
where $\ell_{hk}$ denotes the distance between the 3D points $h$ and $k$, $l_{hi}$ denotes the distance between camera centre $i$ and 3D point $h$, $l_{ki}$ denotes the distance between camera centre $i$ and 3D point $k$, and $\theta_{hik}$ denotes the angle between the camera center and the two points.
Observe that such distances are all unknown. See Figure \ref{fig:SWM_graph}. The angle can be computed as \cite{HartleyZisserman04}:
\begin{equation}
     \cos(\theta_{hik}) = \frac{
\mathbf{m}_{ih}^{\mathsf{T}} \omega_i \ \mathbf{m}_{ik}
}{
\sqrt{\mathbf{m}_{ih}^{\mathsf{T}} \omega_i \ \mathbf{m}_{ih} } \ 
\sqrt{\mathbf{m}_{ik}^{\mathsf{T}} \omega_i \ \mathbf{m}_{ik}}
},
\ \omega_i = (K_i K_i^{\mathsf{T}})^{-1}
\end{equation}
where $\mathbf{m}_{ih}$ and $\mathbf{m}_{ik}$ denotes the projections of 3D points $h$ and $k$ onto image $i$, respectively.
Recall that $K_i$ denotes the calibration matrix of the $i$-th camera, which is known in a calibrated scenario (see Section \ref{subsec_notation}).
Observe that Equation \eqref{eq_law_cosines} involves a single triplet in the graph. When considering the entire graph simultaneously, the problem can be expressed as semidefinite programming, as shown in \cite{Li10}. To summarize, this approach follows two main steps:
\begin{enumerate}
    \item Compute distances between 3D points by solving \eqref{eq_law_cosines};
    \item Recover coordinates of 3D points starting from their mutual distances with standard tools \cite{AspnesErenAl06}.
\end{enumerate}

\begin{remark}
The approach discussed in this section works well provided that, in theory, it is possible to uniquely recover 3D points starting from a subset of pairwise distances (up to rotation, translation and reflection). This happens when the underlying graph is \textbf{rigid} in the classical sense \cite{AspnesErenAl06}. 
Observe that distance-based rigidity is different than parallel rigidity (see Section \ref{subsec_theory_SFM}), and it is applied to another graph here.   
\label{remark_theory_SWM}
\end{remark}

\begin{remark}
The approach reviewed in this section is also known as \textbf{Structure without Motion}, since it has eliminated the need of estimating the motion parameters. Hence, in principle, it is more direct and ``economic''. Observe that this represents a paradigm shift with respect to most SfM approaches, where motion constitutes an essential component. However, the current implementation of \cite{Li10} is computationally expensive, being based on semidefinite programming, thereby limiting the usage of this method in practice. In this respect, some extensions were proposed in the literature, including a simplified formulation with homographies \cite{JiangLinAl15}, and the idea of algebraic variable elimination \cite{ZhangAliagaAl06,AliagaZhangAl07}. 
\end{remark}

\begin{remark} 
We conclude this section by briefly mentioning a related recent approach, MASt3R-SfM \cite{DuisterhofZustAl25}. Although considering a different framework/assumptions than this survey (similarly to the method discussed in Remark \ref{remark_VGGSFM}), it is analogous in principle to the ``structure without motion'' philosophy. Specifically, MASt3R-SfM directly estimates the structure from pairs of images (using a deep network): this results in multiple (independent) two-view reconstructions, which are then brought into a common reference frame, as also done by Hierarchical SfM (see Remark \ref{remark_hierarchical}). The training loss involves both a matching error and the reprojection error.
This method achieves remarkable results on a variety of datasets.
\end{remark}

\section{Conclusion}
\label{sec_conclusion}

This paper has reviewed existing methods and theory related to  Structure from Motion, following a conceptual taxonomy centered on the interaction between the two main parts of the problem, that are motion and structure: 
the first category in the proposed taxonomy considers structure and motion in a joint manner; the second category focuses on motion; finally, the third category focuses on structure.
General advantages and disadvantages of these distinct formulations have been illustrated in order to give practical insights, with the provision that -- in real applications -- the choice of the most suitable method depends on the specific scenario and assumptions. In this context, the usage of deep learning as an aid for traditional approaches/principles demands for further research.

Throughout this paper, special attention has been devoted to illustrate degenerate cases inherent in different SfM formulations, where a unique reconstruction is unattainable. Note that the knowledge of those degenerate configurations is relevant from the theoretical perspective. However, most existing methods do not explicitly check for the presence/absence of degeneracies. 
In other terms, a better synergy is required between theory and practice, that should be addressed in a unified manner: we hope that this survey can serve as a starting point for future SfM methods implementing this idea.

Finally, other related problems not addressed in this paper received limited attention compared to the plain SfM task and are surely worth investigating in the future. Notable examples include SfM with objects \cite{GayRubinoAl17}, SfM with lines \cite{LiuGaoAl24}, privacy-preserving SfM \cite{GeppertLarssonAl20}, and multibody SfM \cite{OzdenSchindlerAl10}.

\section*{Acknowledgements} 
This paper is supported by PNRR-PE-AI FAIR project funded by the NextGeneration EU program. The author would like to thank Andrea Fusiello and Luca Magri for their help and useful feedback on the material prepared for recent tutorials \cite{MagriArrigoni22,MagriArrigoni24}, which set the basis for this paper.


 

\bibliographystyle{IEEEtran}
\bibliography{Definitions,FedeNEW,AndreaNEW,LucaNEW,NostriNEW}

\begin{thebibliography}{100}
\providecommand{\url}[1]{#1}
\csname url@samestyle\endcsname
\providecommand{\newblock}{\relax}
\providecommand{\bibinfo}[2]{#2}
\providecommand{\BIBentrySTDinterwordspacing}{\spaceskip=0pt\relax}
\providecommand{\BIBentryALTinterwordstretchfactor}{4}
\providecommand{\BIBentryALTinterwordspacing}{\spaceskip=\fontdimen2\font plus
\BIBentryALTinterwordstretchfactor\fontdimen3\font minus
  \fontdimen4\font\relax}
\providecommand{\BIBforeignlanguage}[2]{{%
\expandafter\ifx\csname l@#1\endcsname\relax
\typeout{** WARNING: IEEEtran.bst: No hyphenation pattern has been}%
\typeout{** loaded for the language `#1'. Using the pattern for}%
\typeout{** the default language instead.}%
\else
\language=\csname l@#1\endcsname
\fi
#2}}
\providecommand{\BIBdecl}{\relax}
\BIBdecl

\bibitem{OzyesilVoroninskiAl17}
O.~Ozyesil, V.~Voroninski, R.~Basri, and A.~Singer, ``A survey of structure
  from motion,'' \emph{Acta Numerica}, vol.~26, pp. 305 -- 364, 2017.

\bibitem{HartleyZisserman04}
R.~Hartley and A.~Zisserman, \emph{Multiple View Geometry in Computer Vision},
  2nd~ed.\hskip 1em plus 0.5em minus 0.4em\relax Cambridge University Press,
  2004.

\bibitem{CrandallOwensAl11}
D.~Crandall, A.~Owens, N.~Snavely, and D.~P. Huttenlocher,
  ``Discrete-continuous optimization for large-scale structure from motion,''
  in \emph{Proceedings of the IEEE Conference on Computer Vision and Pattern
  Recognition}, 2011, pp. 3001--3008.

\bibitem{ChatterjeeGovindu13}
A.~Chatterjee and V.~M. Govindu, ``Efficient and robust large-scale rotation
  averaging,'' in \emph{Proceedings of the International Conference on Computer
  Vision}, 2013.

\bibitem{WilsonSnavely14}
K.~Wilson and N.~Snavely, ``Robust global translations with {1DSfM},'' in
  \emph{Proceedings of the European Conference on Computer Vision}, 2014, pp.
  61--75.

\bibitem{SenguptaAmirAl17}
S.~{Sengupta}, T.~{Amir}, M.~{Galun}, T.~{Goldstein}, D.~W. {Jacobs},
  A.~{Singer}, and R.~{Basri}, ``A new rank constraint on multi-view
  fundamental matrices, and its application to camera location recovery,'' in
  \emph{Proceedings of the IEEE Conference on Computer Vision and Pattern
  Recognition}, 2017, pp. 2413--2421.

\bibitem{MagerandDel-Bue20}
L.~Magerand and A.~Del~Bue, ``Revisiting projective structure from motion: A
  robust and efficient incremental solution,'' \emph{IEEE Transactions on
  Pattern Analysis and Machine Intelligence}, vol.~42, no.~2, pp. 430--443,
  2020.

\bibitem{SarlinLindenbergerAl23}
P.-E. Sarlin, P.~Lindenberger, V.~Larsson, and M.~Pollefeys, ``Pixel-perfect
  structure-from-motion with featuremetric refinement,'' \emph{IEEE
  Transactions on Pattern Analysis and Machine Intelligence}, 2023.

\bibitem{HartleyTrumpfAl13}
R.~I. Hartley, J.~Trumpf, Y.~Dai, and H.~Li, ``Rotation averaging,''
  \emph{International Journal of Computer Vision}, 2013.

\bibitem{NasihatkonHartleyAl15}
B.~Nasihatkon, R.~Hartley, and J.~Trumpf, ``A generalized projective
  reconstruction theorem and depth constraints for projective factorization,''
  \emph{International Journal of Computer Vision}, vol. 115, pp. 87 -- 115,
  2015.

\bibitem{WilsonBindel20}
K.~{Wilson} and D.~{Bindel}, ``On the distribution of minima in
  intrinsic-metric rotation averaging,'' in \emph{Proceedings of the IEEE
  Conference on Computer Vision and Pattern Recognition}, 2020, pp. 6030--6038.

\bibitem{BratelundRydell23}
M.~Bratelund and F.~Rydell, ``Compatibility of fundamental matrices for
  complete viewing graphs,'' in \emph{Proceedings of the International
  Conference on Computer Vision}, 2023, pp. 3305 -- 3313.

\bibitem{ManamGovindu23}
L.~Manam and V.~M. Govindu, ``Sensitivity in translation averaging,'' in
  \emph{Neural Information Processing Systems (NeurIPS)}, 2023.

\bibitem{KastenGeifmanAl19}
Y.~{Kasten}, A.~{Geifman}, M.~{Galun}, and R.~{Basri}, ``{GPSfM}: Global
  projective {SFM} using algebraic constraints on multi-view fundamental
  matrices,'' in \emph{Proceedings of the IEEE Conference on Computer Vision
  and Pattern Recognition}, 2019, pp. 3259--3267.

\bibitem{MagriArrigoni22}
L.~Magri and F.~Arrigoni, ``Inside plato’s door: a tour in multi-view
  geometry,'' Tutorial -- in conjunction with the Conference on Computer Vision
  and Pattern Recognition (CVPR), 2022,
  \url{https://sites.google.com/view/platomultiview/}.

\bibitem{MagriArrigoni24}
------, ``Inside plato’s door: a tour in multi-view geometry,'' Tutorial --
  in conjunction with the European Conference on Computer Vision (ECCV), 2024,
  \url{https://sites.google.com/view/platomultiview24/home-page}.

\bibitem{HeSunAl24}
X.~He, J.~Sun, Y.~Wang, S.~Peng, Q.~Huang, H.~Bao, and X.~Zhou, ``Detector-free
  structure from motion,'' in \emph{2024 IEEE/CVF Conference on Computer Vision
  and Pattern Recognition (CVPR)}, 2024, pp. 21\,594--21\,603.

\bibitem{SaputraMarkhamAl18}
M.~R.~U. Saputra, A.~Markham, and N.~Trigoni, ``Visual {SLAM} and structure
  from motion in dynamic environments: A survey,'' \emph{ACM Computing
  Surveys}, vol.~51, no.~2, pp. 37:1--37:36, 2018.

\bibitem{FahimAminAl21}
G.~Fahim, K.~Amin, and S.~Zarif, ``Single-view 3d reconstruction: A survey of
  deep learning methods,'' \emph{Computers $\&$ Graphics}, vol.~94, pp.
  164--190, 2021.

\bibitem{Bratelund24b}
M.~Bratelund, ``Critical configurations for two projective views, a new
  approach,'' \emph{Journal of Symbolic Computation}, vol. 120, 2024.

\bibitem{KahlHartley02}
F.~Kahl and R.~Hartley, ``Critical curves and surfaces for euclidean
  reconstruction,'' in \emph{Proceedings of the European Conference on Computer
  Vision}.\hskip 1em plus 0.5em minus 0.4em\relax Springer Berlin Heidelberg,
  2002, pp. 447--462.

\bibitem{KileelKohn23}
J.~Kileel and K.~Kohn, ``Snapshot of algebraic vision,'' \emph{arXiv}, no.
  2210.11443, 2023.

\bibitem{TriggsMcLauchlanAl00}
B.~Triggs, P.~F. McLauchlan, R.~I. Hartley, and A.~W. Fitzgibbon, ``Bundle
  adjustment - a modern synthesis,'' in \emph{Proceedings of the International
  Workshop on Vision Algorithms}.\hskip 1em plus 0.5em minus 0.4em\relax
  Springer-Verlag, 2000, pp. 298--372.

\bibitem{HollandWelsch77}
P.~W. Holland and R.~E. Welsch, ``Robust regression using iteratively
  reweighted least-squares,'' \emph{Communications in Statistics - Theory and
  Methods}, vol.~6, no.~9, pp. 813--827, 1977.

\bibitem{ZhangZhuAl17}
R.~Zhang, S.~Zhu, T.~Fang, and L.~Quan, ``Distributed very large scale bundle
  adjustment by global camera consensus,'' in \emph{2017 IEEE International
  Conference on Computer Vision (ICCV)}, 2017, pp. 29--38.

\bibitem{LeiZixinAl20}
L.~Zhou, Z.~Luo, M.~Zhen, T.~Shen, S.~Li, Z.~Huang, T.~Fang, and L.~Quan,
  ``Stochastic bundle adjustment for efficient and scalable 3d
  reconstruction,'' in \emph{Computer Vision -- ECCV 2020}, A.~Vedaldi,
  H.~Bischof, T.~Brox, and J.-M. Frahm, Eds.\hskip 1em plus 0.5em minus
  0.4em\relax Cham: Springer International Publishing, 2020, pp. 364--379.

\bibitem{RenLiangAl22}
J.~Ren, W.~Liang, R.~Yan, L.~Mai, S.~Liu, and X.~Liu, ``Megba: A gpu-based
  distributed library for large-scale bundle adjustment,'' in \emph{Computer
  Vision -- ECCV 2022}, S.~Avidan, G.~Brostow, M.~Ciss{\'e}, G.~M. Farinella,
  and T.~Hassner, Eds.\hskip 1em plus 0.5em minus 0.4em\relax Cham: Springer
  Nature Switzerland, 2022, pp. 715--731.

\bibitem{ZhengChenAl23}
M.~Zheng, N.~Chen, J.~Zhu, X.~Zeng, H.~Qiu, Y.~Jiang, X.~Lu, and H.~Qu,
  ``Distributed bundle adjustment with block-based sparse matrix compression
  for super large scale datasets,'' in \emph{2023 IEEE/CVF International
  Conference on Computer Vision (ICCV)}, 2023.

\bibitem{OrtizPupilliAl20}
J.~Ortiz, M.~Pupilli, S.~Leutenegger, and A.~J. Davison, ``Bundle adjustment on
  a graph processor,'' \emph{2020 IEEE/CVF Conference on Computer Vision and
  Pattern Recognition (CVPR)}, pp. 2413--2422, 2020.

\bibitem{DemmelSommerAl21}
N.~Demmel, C.~Sommer, D.~Cremers, and V.~C. Usenko, ``Square root bundle
  adjustment for large-scale reconstruction,'' \emph{2021 IEEE/CVF Conference
  on Computer Vision and Pattern Recognition (CVPR)}, pp. 11\,718--11\,727,
  2021.

\bibitem{WeberDemmelAl23}
S.~Weber, N.~Demmel, T.~C. Chan, and D.~Cremers, ``Power bundle adjustment for
  large-scale 3d reconstruction,'' in \emph{2023 IEEE/CVF Conference on
  Computer Vision and Pattern Recognition (CVPR)}, 2023, pp. 281--289.

\bibitem{HongZachAl16}
J.~H. Hong, C.~Zach, A.~Fitzgibbon, and R.~Cipolla, ``Projective bundle
  adjustment from arbitrary initialization using the variable projection
  method,'' in \emph{Computer Vision -- ECCV 2016}.\hskip 1em plus 0.5em minus
  0.4em\relax Springer International Publishing, 2016, pp. 477--493.

\bibitem{ZachHong18}
C.~Zach and J.~H. Hong, ``pose: Pseudo object space error for
  initialization-free bundle adjustment,'' in \emph{2018 IEEE/CVF Conference on
  Computer Vision and Pattern Recognition}, 2018, pp. 1876--1885.

\bibitem{IglesiasNilssonAl23}
J.~P. Iglesias, A.~Nilsson, and C.~Olsson, ``expose: Accurate
  initialization-free projective factorization using exponential
  regularization,'' in \emph{2023 IEEE/CVF Conference on Computer Vision and
  Pattern Recognition (CVPR)}, 2023, pp. 8959--8968.

\bibitem{WeberHongAl24}
S.~Weber, J.~H. Hong, and D.~Cremers, ``Power variable projection
  for initialization-free large-scale bundle adjustment,'' in \emph{Computer
  Vision -- ECCV 2024}.\hskip 1em plus 0.5em minus 0.4em\relax Cham: Springer
  Nature Switzerland, 2025, pp. 111--126.

\bibitem{SnavelySeitzAl06}
N.~Snavely, S.~M. Seitz, and R.~Szeliski, ``Photo tourism: exploring photo
  collections in {3D},'' in \emph{SIGGRAPH: International Conference on
  Computer Graphics and Interactive Techniques}, 2006, pp. 835--846.

\bibitem{AgarwalSnavelyAl09}
S.~Agarwal, N.~Snavely, I.~Simon, S.~M. Seitz, and R.~Szeliski, ``Building rome
  in a day,'' in \emph{IEEE International Conference on Computer Vision}, 2009.

\bibitem{FrahmFite-GeorgelAl10}
J.-M. Frahm, P.~Fite-Georgel, D.~Gallup, T.~Johnson, R.~Raguram, C.~Wu, Y.-H.
  Jen, E.~Dunn, B.~Clipp, S.~Lazebnik, and M.~Pollefeys, ``Building {R}ome on a
  cloudless day,'' in \emph{Proceedings of the 11th European conference on
  Computer vision: Part IV}, 2010, pp. 368--381.

\bibitem{Wu13}
C.~Wu, ``Towards linear-time incremental structure from motion,'' in
  \emph{Proceedings of the International Conference on 3D Vision (3DV)}.\hskip
  1em plus 0.5em minus 0.4em\relax IEEE, 2013.

\bibitem{SchonbergerFrahm16}
J.~L. Schonberger and J.-M. Frahm, ``Structure-from-motion revisited,'' in
  \emph{Proceedings of the IEEE Conference on Computer Vision and Pattern
  Recognition}, 2016, pp. 4104 -- 4113.

\bibitem{ToldoGherardiAl15}
R.~Toldo, R.~Gherardi, M.~Farenzena, and A.~Fusiello, ``Hierarchical
  structure-and-motion recovery from uncalibrated images,'' \emph{Computer
  Vision and Image Understanding}, 2015.

\bibitem{NiDellaert12}
K.~Ni and F.~Dellaert, ``Hypersfm,'' in \emph{Proceedings of the Joint
  3DIM/3DPVT Conference: 3D Imaging, Modeling, Processing, Visualization and
  Transmission}, 2012, pp. 144--151.

\bibitem{BeslMcKay92}
P.~Besl and N.~McKay, ``A method for registration of {3-D} shapes,'' \emph{IEEE
  Transactions on Pattern Analysis and Machine Intelligence}, vol.~14, no.~2,
  pp. 239--256, February 1992.

\bibitem{SturmTriggs96}
P.~Sturm and B.~Triggs, ``A factorization based algorithm for multi-image
  projective structure and motion,'' in \emph{Proceedings of the European
  Conference on Computer Vision}, 1996, pp. 709--720.

\bibitem{OliensisHartley07}
J.~Oliensis and R.~I. Hartley, ``Iterative extensions of the sturm/triggs
  algorithm: Convergence and nonconvergence,'' \emph{IEEE Transactions on
  Pattern Analysis and Machine Intelligence}, vol.~29, no.~12, pp. 2217--2233,
  2007.

\bibitem{MartinecPajdla05}
D.~Martinec and T.~Pajdla, ``3d reconstruction by fitting low-rank matrices
  with missing data,'' in \emph{2005 IEEE Computer Society Conference on
  Computer Vision and Pattern Recognition}, 2005.

\bibitem{JiaMartinez09}
H.~Jia and A.~M. Martinez, ``Low-rank matrix fitting based on subspace
  perturbation analysis with applications to structure from motion,''
  \emph{IEEE Transactions on Pattern Analysis and Machine Intelligence},
  vol.~31, no.~5, pp. 841--854, 2009.

\bibitem{DaiLiHe13}
Y.~Dai, H.~Li, and M.~He, ``Projective multiview structure and motion from
  element-wise factorization,'' \emph{IEEE Transactions on Pattern Analysis and
  Machine Intelligence}, vol.~35, no.~9, pp. 2238--2251, 2013.

\bibitem{NguyenKimAl19}
L.~T. Nguyen, J.~Kim, and B.~Shim, ``Low-rank matrix completion: A contemporary
  survey,'' \emph{IEEE Access}, vol.~7, pp. 94\,215--94\,237, 2019.

\bibitem{MoranKoslowskyAl21}
D.~Moran, H.~Koslowsky, Y.~Kasten, H.~Maron, M.~Galun, and R.~Basri, ``Deep
  permutation equivariant structure from motion,'' in \emph{Proceedings of the
  IEEE/CVF International Conference on Computer Vision (ICCV)}, 2021, pp.
  5976--5986.

\bibitem{KathibKastenAl25}
F.~Khatib1, Y.~Kasten, D.~Moran, M.~Galun, and R.~Basri, ``Resfm: Robust deep
  equivariant structure from motion,'' in \emph{International Conference on
  Learning Representations (ICLR)}, 2025.

\bibitem{BrynteIglesiasAl24}
L.~Brynte, J.~P. Iglesias, C.~Olsson, and F.~Kahl, ``Learning
  structure-from-motion with graph attention networks,'' in \emph{2024 IEEE/CVF
  Conference on Computer Vision and Pattern Recognition (CVPR)}, 2024, pp.
  4808--4817.

\bibitem{Thompson66}
E.~Thompson, ``Space resection: Failure cases,'' \emph{Photogrammetric Record},
  vol.~X, no.~27, pp. 201--204, 1966.

\bibitem{LeviWerman03}
N.~Levi and M.~Werman, ``The viewing graph,'' in \emph{Proceedings of the IEEE
  Conference on Computer Vision and Pattern Recognition}, 2003, pp. 518 -- 522.

\bibitem{ShenZhuAl16}
T.~Shen, S.~Zhu, T.~Fang, R.~Zhang, and L.~Quan, ``Graph-based consistent
  matching for structure-from-motion,'' in \emph{Proceedings of the European
  Conference on Computer Vision}, 2016, pp. 139 -- 155.

\bibitem{ShanChariAl18}
R.~Shah, V.~Chari, and P.~J. Narayanan, ``View-graph selection framework for
  sfm,'' in \emph{Computer Vision -- ECCV 2018}.\hskip 1em plus 0.5em minus
  0.4em\relax Springer International Publishing, 2018, pp. 553--568.

\bibitem{CuiFragosoAl17}
Q.~Cui, V.~Fragoso, C.~Sweeney, and P.~Sen, ``Graphmatch: Efficient large-scale
  graph construction for structure from motion.'' in \emph{International
  Conference on 3D Vision (3DV)}.\hskip 1em plus 0.5em minus 0.4em\relax IEEE
  Computer Society, 2017, pp. 165--174.

\bibitem{LiShiAl24}
S.~Li, Y.~Shi, and G.~Lerman, ``Efficient detection of long consistent cycles
  and its application to distributed synchronization,'' in \emph{2024 IEEE/CVF
  Conference on Computer Vision and Pattern Recognition (CVPR)}, 2024, pp.
  5260--5269.

\bibitem{Govindu06}
V.~M. Govindu, ``Robustness in motion averaging,'' in \emph{Proceedings of the
  Asian Conference on Computer Vision}, 2006, pp. 457--466.

\bibitem{ZachKlopschitzAl10}
C.~Zach, M.~Klopschitz, and M.~Pollefeys, ``Disambiguating visual relations
  using loop constraints,'' in \emph{Proceedings of the IEEE Conference on
  Computer Vision and Pattern Recognition}, 2010, pp. 1426 -- 1433.

\bibitem{BourmaudMegretAl14}
G.~Bourmaud, R.~Megret, A.~Giremus, and Y.~Berthoumieu, ``Global motion
  estimation from relative measurements in the presence of outliers,'' in
  \emph{Proceedings of the Asian Conference on Computer Vision}, 2014.

\bibitem{ManamGovindu24}
L.~Manam and V.~M. Govindu, ``Leveraging camera triplets for efficient and
  accurate structure-from-motion,'' in \emph{Proceedings of the IEEE Conference
  on Computer Vision and Pattern Recognition}, 2024.

\bibitem{Singer11}
A.~Singer, ``Angular synchronization by eigenvectors and semidefinite
  programming,'' \emph{Applied and Computational Harmonic Analysis}, vol.~30,
  no.~1, pp. 20 -- 36, 2011.

\bibitem{Arie-NachimsonKovalskyAl12}
M.~Arie-Nachimson, S.~Z. Kovalsky, I.~Kemelmacher-Shlizerman, A.~Singer, and
  R.~Basri, ``Global motion estimation from point matches,'' \emph{Proceedings
  of the Joint 3DIM/3DPVT Conference: 3D Imaging, Modeling, Processing,
  Visualization and Transmission}, 2012.

\bibitem{MartinecPajdla07}
D.~Martinec and T.~Pajdla, ``Robust rotation and translation estimation in
  multiview reconstruction,'' in \emph{Proceedings of the IEEE Conference on
  Computer Vision and Pattern Recognition}, 2007.

\bibitem{ArrigoniRossiAl16}
F.~Arrigoni, B.~Rossi, and A.~Fusiello, ``Spectral synchronization of multiple
  views in {SE(3)},'' \emph{SIAM Journal on Imaging Sciences}, vol.~9, no.~4,
  pp. 1963 -- 1990, 2016.

\bibitem{ArrigoniRossiAl18}
F.~Arrigoni, B.~Rossi, P.~Fragneto, and A.~Fusiello, ``Robust synchronization
  in {SO(3)} and {SE(3)} via low-rank and sparse matrix decomposition,''
  \emph{Computer Vision and Image Understanding}, vol. 174, pp. 95--113, 2018.

\bibitem{TejusZaraAl23}
G.~Tejus, G.~Zara, P.~Rota, A.~Fusiello, E.~Ricci, and F.~Arrigoni, ``Rotation
  synchronization via deep matrix factorization,'' in \emph{2023 IEEE
  International Conference on Robotics and Automation (ICRA)}, 2023.

\bibitem{Govindu01}
V.~M. Govindu, ``Combining two-view constraints for motion estimation,'' in
  \emph{Proceedings of the IEEE Conference on Computer Vision and Pattern
  Recognition}, 2001.

\bibitem{ShiLerman20}
Y.~Shi and G.~Lerman, ``Message passing least squares framework and its
  application to rotation synchronization,'' in \emph{Proceedings of the 37th
  International Conference on Machine Learning}, ser. Proceedings of Machine
  Learning Research, vol. 119.\hskip 1em plus 0.5em minus 0.4em\relax PMLR,
  2020, pp. 8796--8806.

\bibitem{ErikssonOlssonAl18}
A.~{Eriksson}, C.~{Olsson}, F.~{Kahl}, and T.-J. {Chin}, ``Rotation averaging
  and strong duality,'' in \emph{Proceedings of the IEEE Conference on Computer
  Vision and Pattern Recognition}, 2018, pp. 127--135.

\bibitem{DellaertRosenAl20}
F.~Dellaert, D.~M. Rosen, J.~Wu, R.~Mahony, and L.~Carlone, ``Shonan rotation
  averaging: Global optimality by surfing $so(p)^n$,'' in \emph{Computer Vision
  -- ECCV 2020}.\hskip 1em plus 0.5em minus 0.4em\relax Springer International
  Publishing, 2020, pp. 292--308.

\bibitem{ParraChngAl21}
A.~Parra, S.-F. Chng, T.-J. Chin, A.~Eriksson, and I.~Reid, ``Rotation
  coordinate descent for fast globally optimal rotation averaging,'' in
  \emph{Proceedings of the IEEE/CVF Conference on Computer Vision and Pattern
  Recognition (CVPR)}, 2021, pp. 4298--4307.

\bibitem{MoreiraMarquesAl21}
G.~Moreira, M.~Marques, and J.~a.~P. Costeira, ``Rotation averaging in a split
  second: A primal-dual method and a closed-form for cycle graphs,'' in
  \emph{Proceedings of the IEEE/CVF International Conference on Computer Vision
  (ICCV)}, 2021, pp. 5452--5460.

\bibitem{BirdalArbelAl20}
T.~{Birdal}, M.~{Arbel}, U.~{Simsekli}, and L.~J. {Guibas}, ``Synchronizing
  probability measures on rotations via optimal transport,'' in
  \emph{Proceedings of the IEEE Conference on Computer Vision and Pattern
  Recognition}, 2020, pp. 1566--1576.

\bibitem{ZhangLarssonAl23}
G.~Zhang, V.~Larsson, and D.~Barath, ``Revisiting rotation averaging:
  Uncertainties and robust losses,'' in \emph{Proceedings of the IEEE
  Conference on Computer Vision and Pattern Recognition}, 2023.

\bibitem{PanPollefeysAl24}
L.~Pan, M.~Pollefeys, and D.~Bar{\'a}th, ``Gravity-aligned rotation averaging
  with circular regression,'' in \emph{Computer Vision -- ECCV 2024}.\hskip 1em
  plus 0.5em minus 0.4em\relax Cham: Springer Nature Switzerland, 2025, pp.
  97--116.

\bibitem{LeeCivera22}
S.~H. Lee and J.~Civera, ``Hara: A hierarchical approach for robust rotation
  averaging,'' in \emph{2022 IEEE/CVF Conference on Computer Vision and Pattern
  Recognition (CVPR)}, 2022.

\bibitem{PurkaitChinAl20}
P.~Purkait, T.-J. Chin, and I.~Reid, ``Neurora: Neural robust rotation
  averaging,'' in \emph{Computer Vision -- ECCV 2020}.\hskip 1em plus 0.5em
  minus 0.4em\relax Springer International Publishing, 2020, pp. 137--154.

\bibitem{YangLiAl21}
L.~Yang, H.~Li, J.~A. Rahim, Z.~Cui, and P.~Tan, ``End-to-end rotation
  averaging with multi-source propagation,'' in \emph{Proceedings of the
  IEEE/CVF Conference on Computer Vision and Pattern Recognition (CVPR)}, June
  2021, pp. 11\,774--11\,783.

\bibitem{LiLing21}
X.~Li and H.~Ling, ``Pogo-net: Pose graph optimization with graph neural
  networks,'' in \emph{Proceedings of the IEEE/CVF International Conference on
  Computer Vision (ICCV)}, October 2021, pp. 5895--5905.

\bibitem{LiCuiAl22}
H.~Li, Z.~Cui, S.~Liu, and P.~Tan, ``Rago: Recurrent graph optimizer for
  multiple rotation averaging,'' in \emph{2022 IEEE/CVF Conference on Computer
  Vision and Pattern Recognition (CVPR)}, 2022.

\bibitem{HartleyAftabAl11}
R.~Hartley, K.~Aftab, and J.~Trumpf, ``L1 rotation averaging using the
  {Weiszfeld} algorithm,'' \emph{Proceedings of the IEEE Conference on Computer
  Vision and Pattern Recognition}, pp. 3041--3048, 2011.

\bibitem{TronZhouAl16}
R.~Tron, X.~Zhou, and K.~Daniilidis, ``A survey on rotation optimization in
  structure from motion,'' in \emph{Computer Vision and Pattern Recognition
  Workshops (CVPRW)}, 2016.

\bibitem{BoumalSingerAl14}
N.~Boumal, A.~Singer, P.~A. Absil, and V.~D. Blondel, ``{Cramer-Rao} bounds for
  synchronization of rotations,'' \emph{Information and Inference: A Journal of
  the IMA}, vol.~3, no.~1, pp. 1 -- 39, 2014.

\bibitem{WilsonBindelAl16}
K.~Wilson, D.~Bindel, and N.~Snavely, ``When is rotations averaging hard?'' in
  \emph{Proceedings of the European Conference on Computer Vision}, 2016, pp.
  255 -- 270.

\bibitem{BhowmickPatraAl14}
B.~Bhowmick, S.~Patra, A.~Chatterjee, V.~M. Govindu, and S.~Banerjee, ``Divide
  and conquer: Efficient large-scale structure from motion using graph
  partitioning,'' in \emph{12th Asian Conference on Computer Vision (ACCV
  2014)}, 2014.

\bibitem{DalcinMagriAl21}
A.~{Porfiri Dal Cin}, L.~Magri, F.~Arrigoni, A.~Fusiello, and G.~Boracchi,
  ``Synchronization of group-labelled multi-graphs,'' in \emph{Proceedings of
  the International Conference on Computer Vision}, 2021.

\bibitem{ZhuZhangAl18}
S.~Zhu, R.~Zhang, L.~Zhou, T.~Shen, T.~Fang, P.~Tan, and L.~Quan, ``Very
  large-scale global sfm by distributed motion averaging,'' in
  \emph{Proceedings of the IEEE Conference on Computer Vision and Pattern
  Recognition}, 2018, pp. 4568--4577.

\bibitem{BrandAntoneAl04}
M.~Brand, M.~Antone, and S.~Teller, ``Spectral solution of large-scale
  extrinsic camera calibration as a graph embedding problem,'' in
  \emph{Proceedings of the European Conference on Computer Vision}, 2004.

\bibitem{OzyesilSingerAl15}
O.~Ozyesil, A.~Singer, and R.~Basri, ``Stable camera motion estimation using
  convex programming,'' \emph{SIAM Journal on Imaging Sciences}, vol.~8, no.~2,
  pp. 1220 -- 1262, 2015.

\bibitem{OzyesilSinger15}
O.~Ozyesil and A.~Singer, ``Robust camera location estimation by convex
  programming,'' in \emph{Proceedings of the IEEE Conference on Computer Vision
  and Pattern Recognition}, 2015, pp. 2674 -- 2683.

\bibitem{HeRuanAl25}
Z.~He, H.~Ruan, and Q.~Huang, ``A robust translation synchronization
  algorithm,'' in \emph{Proceedings of the International Conference on 3D
  Vision (3DV)}, 2025.

\bibitem{ZhuangCheongAl18}
B.~Zhuang, L.-F. Cheong, and G.~H. Lee, ``Baseline desensitizing in translation
  averaging,'' in \emph{2018 IEEE/CVF Conference on Computer Vision and Pattern
  Recognition}, 2018, pp. 4539--4547.

\bibitem{JiangCuiAl13}
N.~Jiang, Z.~Cui, and P.~Tan, ``A global linear method for camera pose
  registration,'' in \emph{Proceedings of the International Conference on
  Computer Vision}, 2013.

\bibitem{MoulonMonasseAl13}
P.~Moulon, P.~Monasse, and R.~Marlet, ``Global fusion of relative motions for
  robust, accurate and scalable structure from motion,'' in \emph{Proceedings
  of the International Conference on Computer Vision}, 2013, pp. 3248--3255.

\bibitem{TronVidal14}
R.~Tron and R.~Vidal, ``Distributed {3-D} localization of camera sensor
  networks from {2-D} image measurements,'' \emph{IEEE Transactions on
  Automatic Control}, vol.~59, no.~12, pp. 3325--3340, 2014.

\bibitem{GoldsteinHandAl16}
T.~Goldstein, P.~Hand, C.~Lee, V.~Voroninski, and S.~Soatto, ``{ShapeFit} and
  {ShapeKick} for robust, scalable structure from motion,'' in
  \emph{Proceedings of the European Conference on Computer Vision}, 2016, pp.
  289 -- 304.

\bibitem{SidharthaGovindu24}
S.~Chitturi and V.~M. Govindu, ``Adaptive annealing for robust averaging,'' in
  \emph{Computer Vision -- ECCV 2024}.\hskip 1em plus 0.5em minus 0.4em\relax
  Cham: Springer Nature Switzerland, 2025, pp. 53--69.

\bibitem{ManamGovindu22}
L.~Manam and V.~M. Govindu, ``Correspondence reweighted translation
  averaging,'' in \emph{Computer Vision -- ECCV 2022}.\hskip 1em plus 0.5em
  minus 0.4em\relax Cham: Springer Nature Switzerland, 2022, pp. 56--72.

\bibitem{ShiLerman18}
Y.~Shi and G.~Lerman, ``Estimation of camera locations in highly corrupted
  scenarios: All about that base, no shape trouble,'' in \emph{2018 IEEE/CVF
  Conference on Computer Vision and Pattern Recognition}, 2018, pp. 2868--2876.

\bibitem{ArrigoniFusielloAl15}
F.~Arrigoni, A.~Fusiello, and B.~Rossi, ``On computing the translations norm in
  the epipolar graph,'' in \emph{Proceedings of the International Conference on
  3D Vision (3DV)}, 2015, pp. 300--308.

\bibitem{KastenGeifmanAl19b}
Y.~Kasten, A.~Geifman, M.~Galun, and R.~Basri, ``Algebraic characterization of
  essential matrices and their averaging in multiview settings,'' in \emph{2019
  IEEE/CVF International Conference on Computer Vision (ICCV)}, 2019, pp.
  5894--5902.

\bibitem{TaoCuiAl24}
P.~Tao, H.~Cui, M.~Rong, and S.~Shen, ``Revisiting global translation
  estimation with feature tracks,'' in \emph{2024 IEEE/CVF Conference on
  Computer Vision and Pattern Recognition (CVPR)}, 2024, pp. 20\,686--20\,696.

\bibitem{PanBarathAl24}
L.~Pan, D.~Barath, M.~Pollefeys, and J.~L. Schonberger, ``Global
  structure-from-motion revisited,'' in \emph{Proceedings of the European
  Conference on Computer Vision}, 2024.

\bibitem{WangKaraevAl24}
J.~Wang, N.~Karaev, C.~Rupprecht, and D.~Novotny, ``Vggsfm: Visual geometry
  grounded deep structure from motion,'' in \emph{2024 IEEE/CVF Conference on
  Computer Vision and Pattern Recognition (CVPR)}, 2024, pp. 21\,686--21\,697.

\bibitem{MadhavanFusielloAl24}
R.~Madhavan, A.~Fusiello, and F.~Arrigoni, ``Synchronization of projective
  transformations,'' in \emph{Proceedings of the European Conference on
  Computer Vision}, 2024.

\bibitem{SinhaPollefeysAl04}
S.~Sinha, M.~Pollefeys, and L.~McMillan, ``Camera network calibration from
  dynamic silhouettes,'' in \emph{Proceedings of the IEEE Conference on
  Computer Vision and Pattern Recognition}, 2004, pp. I--I.

\bibitem{ColomboFanfani21}
C.~Colombo and M.~Fanfani, ``A closed form solution for viewing graph
  construction in uncalibrated vision,'' in \emph{2021 IEEE/CVF International
  Conference on Computer Vision Workshops (ICCVW)}, 2021.

\bibitem{Harary72}
F.~Harary, \emph{Graph Theory}.\hskip 1em plus 0.5em minus 0.4em\relax
  Addison-Wesley, 1972.

\bibitem{ZhaoZelazo16}
S.~{Zhao} and D.~{Zelazo}, ``Localizability and distributed protocols for
  bearing-based network localization in arbitrary dimensions,''
  \emph{Automatica}, vol.~69, pp. 334 -- 341, 2016.

\bibitem{ArrigoniFusiello18}
F.~Arrigoni and A.~Fusiello, ``Bearing-based network localizability: A unifying
  view,'' \emph{IEEE Transactions on Pattern Analysis and Machine
  Intelligence}, vol.~41, no.~9, pp. 2049 -- 2069, 2019.

\bibitem{TronCarloneAl15}
R.~Tron, L.~Carlone, F.~Dellaert, and K.~Daniilidis, ``Rigid components
  identification and rigidity enforcement in bearing-only localization using
  the graph cycle basis,'' in \emph{IEEE American Control Conference}, 2015,
  pp. 3911--3918.

\bibitem{KennedyDaniilidisAl12}
R.~Kennedy, K.~Daniilidis, O.~Naroditsky, and C.~J. Taylor, ``Identifying
  maximal rigid components in bearing-based localization,'' in
  \emph{Proceedings of the International Conference on Intelligent Robots and
  Systems}, 2012, pp. 194 -- 201.

\bibitem{TragerHebertAl15}
M.~{Trager}, M.~{Hebert}, and J.~{Ponce}, ``The joint image handbook,'' in
  \emph{Proceedings of the International Conference on Computer Vision}, 2015,
  pp. 909--917.

\bibitem{ArrigoniFusielloAl24}
F.~Arrigoni, A.~Fusiello, and T.~Pajdla, ``A direct approach to viewing graph
  solvability,'' in \emph{Proceedings of the European Conference on Computer
  Vision}, 2024.

\bibitem{TragerOssermanAl18}
M.~Trager, B.~Osserman, and J.~Ponce, ``On the solvability of viewing graphs,''
  in \emph{Proceedings of the European Conference on Computer Vision}, 2018,
  pp. 335--350.

\bibitem{ArrigoniFusielloAl21}
F.~Arrigoni, A.~Fusiello, E.~Ricci, and T.~Pajdla, ``Viewing graph solvability
  via cycle consistency,'' in \emph{Proceedings of the International Conference
  on Computer Vision}, 2021, pp. 5540 -- 5549.

\bibitem{Dube90}
T.~W. Dub\'e, ``The structure of polynomial ideals and {Gr\"obner} bases,''
  \emph{SIAM Journal on Computing}, vol.~19, no.~4, pp. 750 -- 773, 1990.

\bibitem{ArrigoniPajdlaAl23}
F.~Arrigoni, T.~Pajdla, and A.~Fusiello, ``Viewing graph solvability in
  practice,'' in \emph{Proceedings of the International Conference on Computer
  Vision}, 2023, pp. 8147--8155.

\bibitem{RudiPizzoliAl11}
A.~Rudi, M.~Pizzoli, and F.~Pirri, ``Linear solvability in the viewing graph,''
  in \emph{Proceedings of the Asian Conference on Computer Vision}, 2011, pp.
  369--381.

\bibitem{ArrigoniFusielloAl22}
F.~Arrigoni, A.~Fusiello, R.~Rizzi, E.~Ricci, and T.~Pajdla, ``Revisiting
  viewing graph solvability: an effective approach based on cycle
  consistency,'' \emph{IEEE Transactions on Pattern Analysis and Machine
  Intelligence}, pp. 1--14, 2022.

\bibitem{Li10}
H.~Li, ``Multi-view structure computation without explicitly estimating
  motion,'' in \emph{2010 IEEE Computer Society Conference on Computer Vision
  and Pattern Recognition}, 2010, pp. 2777--2784.

\bibitem{AspnesErenAl06}
J.~Aspnes, T.~Eren, D.~Goldenberg, A.~Morse, W.~Whiteley, Y.~Yang, B.~Anderson,
  and P.~Belhumeur, ``A theory of network localization,'' \emph{IEEE
  Transactions on Mobile Computing}, vol.~5, no.~12, pp. 1663 -- 1678, 2006.

\bibitem{JiangLinAl15}
N.~Jiang, W.-Y. Lin, M.~N. Do, and J.~Lu, ``Direct structure estimation for 3d
  reconstruction,'' in \emph{2015 IEEE Conference on Computer Vision and
  Pattern Recognition (CVPR)}, 2015, pp. 2655--2663.

\bibitem{ZhangAliagaAl06}
J.~Zhang, D.~G. Aliaga, M.~Boutin, and R.~Insley, ``Angle independent bundle
  adjustment refinement,'' in \emph{Third International Symposium on 3D Data
  Processing, Visualization, and Transmission (3DPVT'06)}, 2006, pp.
  1108--1116.

\bibitem{AliagaZhangAl07}
D.~G. Aliaga, J.~Zhang, and M.~Boutin, ``Simplifying the reconstruction of 3d
  models using parameter elimination,'' in \emph{2007 IEEE 11th International
  Conference on Computer Vision}, 2007.

\bibitem{DuisterhofZustAl25}
B.~P. Duisterhof, L.~Zust, P.~Weinzaepfel, V.~Leroy, Y.~Cabon, and J.~Revaud,
  ``Mast3r-sfm: a fully-integrated solution for unconstrained
  structure-from-motion,'' in \emph{International Conference on 3D Vision
  (3DV)}, 2025.

\bibitem{GayRubinoAl17}
P.~Gay, C.~Rubino, V.~Bansal, and A.~Del~Bue, ``Probabilistic structure from
  motion with objects (psfmo),'' in \emph{Proceedings of the IEEE International
  Conference on Computer Vision (ICCV)}, Oct 2017.

\bibitem{LiuGaoAl24}
S.~Liu, Y.~Gao, T.~Zhang, R.~Pautrat, J.~L. Sch{\"o}nberger, V.~Larsson, and
  M.~Pollefeys, ``Robust incremental structure-from-motion with hybrid
  features,'' in \emph{Computer Vision -- ECCV 2024}.\hskip 1em plus 0.5em
  minus 0.4em\relax Cham: Springer Nature Switzerland, 2025, pp. 249--269.

\bibitem{GeppertLarssonAl20}
M.~Geppert, V.~Larsson, P.~Speciale, J.~L. Sch{\"o}nberger, and M.~Pollefeys,
  ``Privacy preserving structure-from-motion,'' in \emph{Computer Vision --
  ECCV 2020}.\hskip 1em plus 0.5em minus 0.4em\relax Cham: Springer
  International Publishing, 2020, pp. 333--350.

\bibitem{OzdenSchindlerAl10}
K.~E. Ozden, K.~Schindler, and L.~Van~Gool, ``Multibody structure-from-motion
  in practice,'' \emph{IEEE Transactions on Pattern Analysis and Machine
  Intelligence}, vol.~32, no.~6, pp. 1134--1141, 2010.

\end{thebibliography}

%

\vfill

\end{document}